\documentclass[sigconf]{aamas} 

\usepackage{graphicx}
\usepackage{subcaption}

\usepackage{booktabs}
\usepackage{multirow}

\usepackage{balance}

\usepackage{bbm}
\usepackage{amsmath}
\usepackage{amsthm}
\usepackage{mathtools}
\usepackage{wrapfig}

\theoremstyle{definition}

\usepackage{algorithm}
\usepackage{algpseudocode}

\newcommand{\ra}{\rangle}
\newcommand{\la}{\langle}

\newcommand{\argmax}{\mathop{\mathrm{arg\,max}}}

\DeclarePairedDelimiterX{\infdivx}[2]{(}{)}{%
  #1\;\delimsize\|\;#2%
}

\usepackage{balance} 

\setcopyright{ifaamas}
\acmConference[AAMAS '21]{Proc.\@ of the 20th International Conference on Autonomous Agents and Multiagent Systems (AAMAS 2021)}{May 3--7, 2021}{Online}{U.~Endriss, A.~Now\'{e}, F.~Dignum, A.~Lomuscio (eds.)}
\copyrightyear{2021}
\acmYear{2021}
\acmDOI{}
\acmPrice{}
\acmISBN{}

\acmSubmissionID{647}

\title[Action Priors for Large Action Spaces in Robotics]{Action Priors for Large Action Spaces in Robotics}

\author{Ondrej Biza}
\affiliation{
  \institution{Northeastern University}
  \city{Boston, MA, USA}}
\email{biza.o@northeastern.edu}

\author{Dian Wang}
\affiliation{
  \institution{Northeastern University}
  \city{Boston, MA, USA}}
\email{wang.dian@northeastern.edu}

\author{Robert Platt}
\thanks{The 3rd, 4th, and 5th authors are listed in alphabetical order and contributed equally. The authors thank Yunus Terzioglu, Tarik Kelestemur, and our anonymous reviewers for helpful feedback. This work was supported by the Intel Corporation, the 3M Corporation, National  Science  Foundation  (1724257, 1724191, 1763878, 1750649, 1835309), NASA (80NSSC19K1474), startup funds from Northeastern University, the Air Force Research Laboratory (AFRL), and DARPA.}
\affiliation{
  \institution{Northeastern University}
  \city{Boston, MA, USA}}
\email{rplatt@ccs.neu.edu}

\author{Jan-Willem van de Meent}
\affiliation{
  \institution{Northeastern University}
  \city{Boston, MA, USA}}
\email{j.vandemeent@northeastern.edu}

\author{Lawson L.S. Wong}
\affiliation{
  \institution{Northeastern University}
  \city{Boston, MA, USA}}
\email{lsw@ccs.neu.edu}

\begin{abstract}
In robotics, it is often not possible to learn useful policies using pure
model-free reinforcement learning without significant reward shaping or
curriculum learning. As a consequence, many researchers rely on expert
demonstrations to guide learning. However, acquiring expert
demonstrations can be expensive. This paper proposes an alternative
approach where the solutions of previously solved tasks are used to
produce an action prior that can facilitate exploration in future tasks.
The action prior is a probability distribution over actions that
summarizes the set of policies found solving previous tasks. Our results
indicate that this approach can be used to solve robotic manipulation
problems that would otherwise be infeasible without expert demonstrations. Source code is available at \url{https://github.com/ondrejba/action_priors}.
\end{abstract}

\keywords{reinforcement learning; deep learning; action prior; robotics; robotic manipulation}

\newcommand{\BibTeX}{\rm B\kern-.05em{\sc i\kern-.025em b}\kern-.08em\TeX}

\begin{document}


\pagestyle{fancy}
\fancyhead{}


\maketitle 


\section{Introduction}

Advances in deep learning have made model-free robot control a viable alternative to model-based motion planning \cite{watter15,finn17deep,levine18}. However, the complexity of tasks solvable by these approaches without extra supervision is limited, partly due to sample inefficiency of deep learning. Hand-crafted temporal abstraction of end-to-end motions such as picking, placing and pushing are a compelling alternative, as they allow agents to reason over longer timescales \cite{zeng18robotic,zeng18learning,platt19}. In particular, \citet{zeng18robotic} proposed a pixel-wise parameterization of the action space, where each pixel in the observed image of the workspace corresponds to a reaching action to that position followed by a pick or place.


While both low-level action spaces with long time horizons and pixel-wise action spaces are difficult to explore, the pixel-wise parameterization makes this challenge more explicit: the agent is presented with thousands of possible actions, and usually only a handful of them enable the agent to make progress toward its goal. Exploration challenges like this are often addressed using reward shaping, curriculum learning, or imitation learning. However, these methods require additional supervision that maybe difficult to provide. Ideally, our agent could learn new skills without expert supervision.

In this paper, we construct priors over the action space -- \emph{action priors} -- that inform the agent of actions that were useful in the context of previously learned tasks. The idea of action priors has existed for some time. \citet{sherstov05} considered a single action prior for all states; later, action priors were extended to state-specific priors \cite{fernandez06,rosman12,rosman15,abel15}. However, to date, action priors have not been applied outside of learning in grid-world-like environments with small action spaces \cite{sherstov05,fernandez06,rosman12,rosman15} and planning in factored models \cite{abel15}.

In contrast, we train action priors in environments with image states and pixel-wise action spaces with thousands of actions. To that end, we represent an action prior as a single fully-convolutional neural network trained to summarize a library of pre-trained policies. We distinguish between a set of training tasks, which we solve using imitation learning, and a held-out set of testing tasks to be solved without expert information. The role of action priors is to bias exploration on the testing tasks toward actions that were found to be useful when solving the training tasks.

\begin{figure*}[t!]
    \centering

    \begin{subfigure}[t]{0.24\textwidth}
        \centering
        \includegraphics[width=1.0\textwidth]{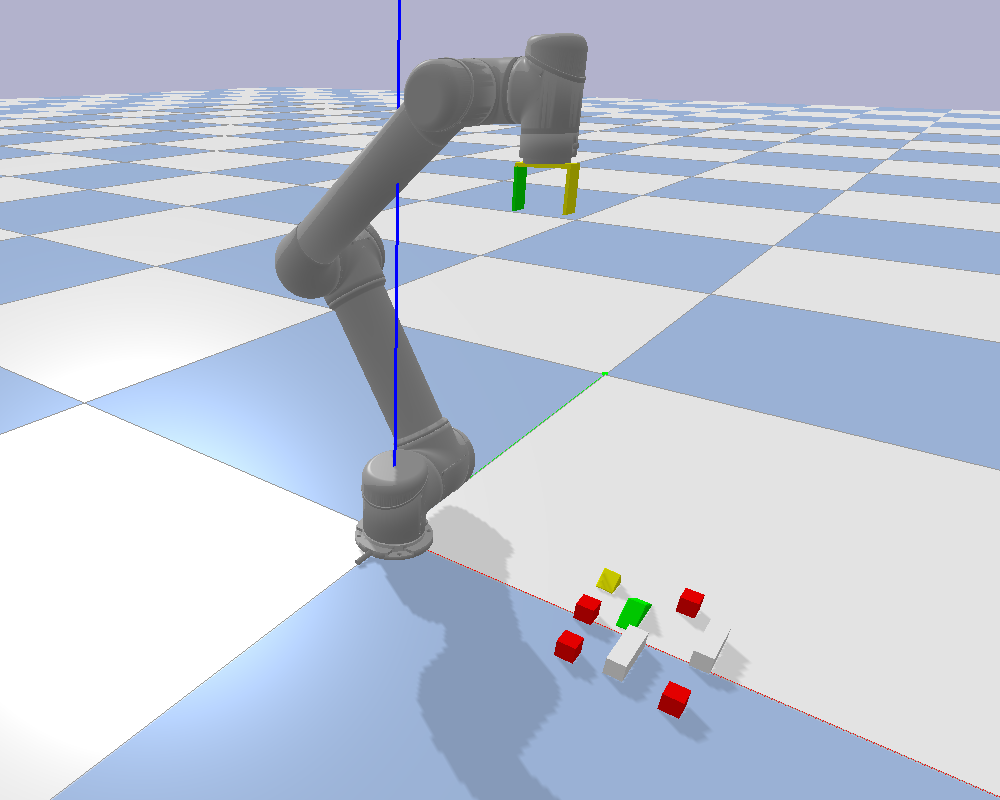}
        \caption{}
    \end{subfigure}
    \begin{subfigure}[t]{0.24\textwidth}
        \centering
        \includegraphics[width=0.8\textwidth]{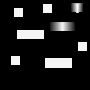}
        \caption{}
    \end{subfigure}
    \begin{subfigure}[t]{0.24\textwidth}
        \centering
        \includegraphics[width=1.0\textwidth]{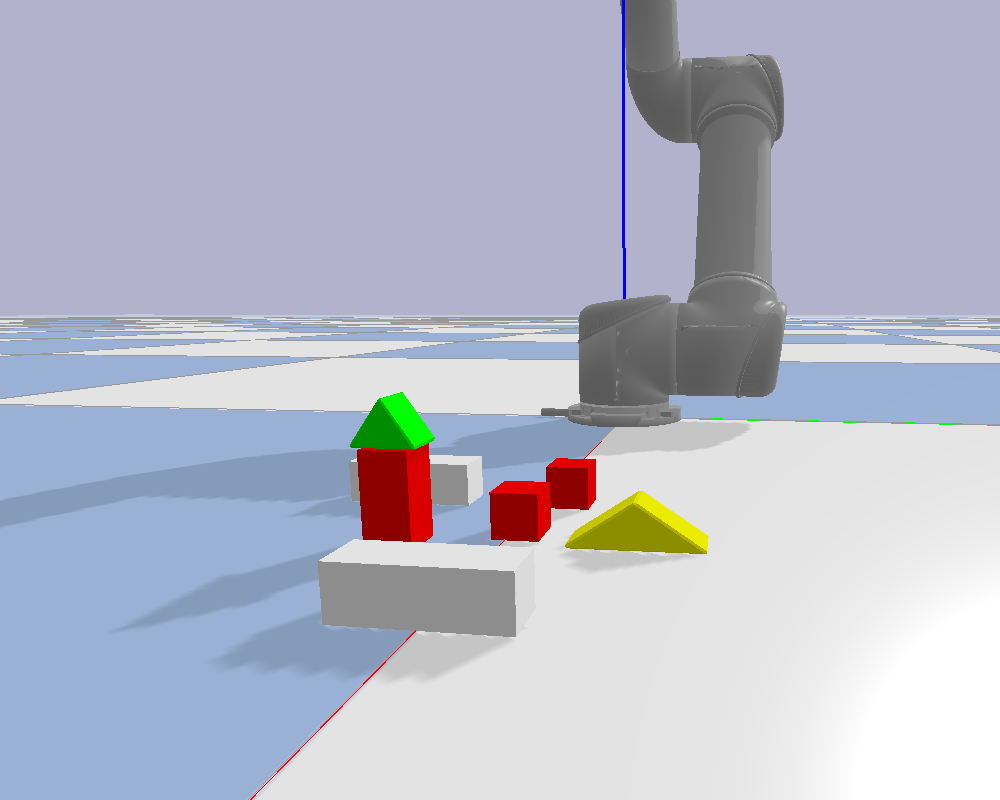}
        \caption{}
    \end{subfigure}
    \begin{subfigure}[t]{0.24\textwidth}
        \centering
        \includegraphics[width=1.0\textwidth]{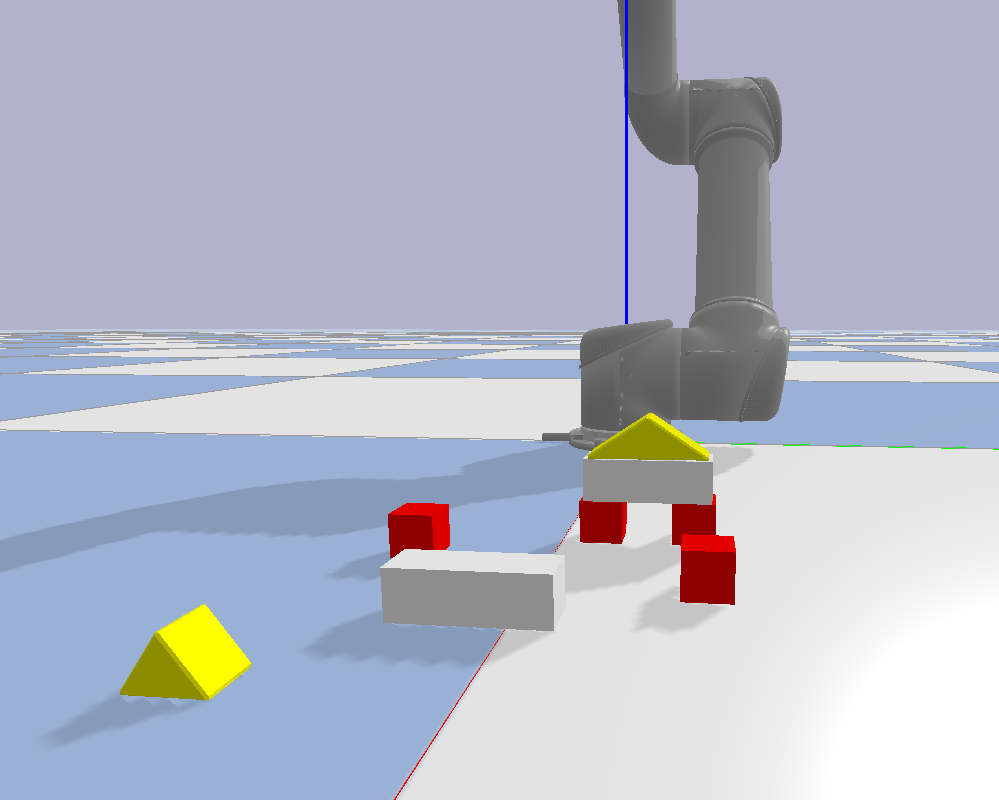}
        \caption{}
    \end{subfigure}

    \caption{Our PyBullet block stacking setup. (a) a simulated UR5 arm and a $60{\times}60$ cm workspace with blocks, (b) a simulated top-down depth camera image of the workspace, (c) and (d) are examples of the goals states of 2 of our 16 block stacking tasks. }
    \label{fig:block_picking_setup}
\end{figure*}

We evaluate our approach on 16 robotic block stacking tasks. We proceed in three stages. First, our agent uses imitation learning to find near-optimal solutions to a subset of the 16 tasks. Second, we condense these near-optimal policies onto a state-dependent probability distribution over actions (i.e. the action prior) that gives high probability to any action that was part of one of the near-optimal policies. Finally, we use the action prior to bias exploration when solving a new task. In the block stacking domain, this action prior gives a high probability to picking actions that are likely to lift a block of some type or placing actions that are likely to result in a stable placement. Although we explicitly focus on robotic manipulation, our approach should generalize well to any problem in robotics with a large action space.

This paper makes two main contributions. First, we show that a state-conditioned action prior is an effective way to transfer knowledge from previously solved tasks to new tasks in a robotic manipulation domain. Our experimental results indicate that this approach can dramatically increase the probability of visiting a goal state during exploration. Second, we introduce a method of learning a state-conditioned action prior in situations where the previously learned policies are valid over different regions of the state/action space. This problem only occurs in large state/action spaces such as in robotic manipulation, and we believe we are the first to address it.

\section{Related Work}
\label{sec:related}

\textbf{Action priors} bias action selection during the exploration phase of learning towards actions that were previously determined to be viable. This information can either be specified by an expert \cite{abel15} or extracted from policies from previously solved tasks \cite{sherstov05,fernandez06,rosman12,rosman15,abel15}. Note that action priors refer to a different construct than policy priors \cite{doshivelez10,wingate11}, as action priors do not involve posterior inference of policy parameters. 

\citet{sherstov05} eliminated actions not optimal for any previous task in a state-agnostic way. Together with their transfer learning algorithm, the state-agnostic action prior increases learning speed in grid-world mazes. \citet{fernandez06,rosman12,rosman15} explored state-specific action priors in similar discrete-state-space MDPs. \citet{fernandez06} alternated between rolling out the policy being learned and a policy sampled from a library. The probability distribution over the library of policies was updated online to maximize rewards for the current task. \citet{rosman12,rosman15} filled in the pseudo-counts of Dirichlet distributions used to select actions in each state by a weighted-sum of actions selected by previously learned policies. \citet{abel15} combined action priors and hand-crafted object-oriented representations \cite{diuk08} to improve the run time of dynamic programming policy search for a Minecraft environment and a real-world robotic manipulation task.

In contrast to pre-defined factored representations in \citet{abel15}, we learn the action prior and model-free policies from pixels. The Dirichlet prior \cite{rosman12,rosman15} is not easily extensible to continuous state spaces; instead, we learn the action prior as a convolutional network. Compared to \citet{fernandez06}, we cannot keep a library of policies loaded in memory, as each policy is parameterized by a large convolutional network--we distill all policies into a single action prior network.

In concurrent work, \citet{ajay20,pertsch20} learned action priors over fixed-length sequences of actions (also called skill priors). Both approaches use varitional autoencoders to learn representations for action sequences, and can be used to solve composite robotic manipulation tasks. \citet{singh20} studied action priors (here called behavior priors) in a setting where training and testing tasks differ in terms of the objects being manipulated, but are otherwise the same.

The topic of \textbf{efficient exploration} is closely related to action prior. Methods in this category often do not use additional information, such as prior policies. Instead, they use a notion of surprise or information content of a visited state. These quantities can be measured by counting the number of times states were visited \cite{strehl05} or by model-based approaches \cite{houthooft16,pathak17}. Our problem statement is incomparable with these approaches, as we exploit additional information from previously learned tasks, which facilitates much more targeted exploration compared to the notion of surprise alone.

\textbf{Transfer learning} has been studied extensively both in classical reinforcement learning \cite{taylor09} and in deep reinforcement learning \cite{parisotto16,teh17,goyal19}. \citet{teh17,goyal19} learned a so-called default policy while learning multiple specialized policies in a multi-task or a multi-goal RL. To transfer to new tasks, \citet{teh17} used the KL-divergence between the default policy and a new policy as regularization, and \citet{goyal19} used their default policy to quantify the notion of a "decision state": a state in which we need make a decision based on the task we want to solve (e.g., a crossroads in a maze). Their agent is then encouraged to explore decision states by adding an intrinsic reward. Both \citet{teh17,goyal19} focuses their experimental evaluation on navigation tasks, with the latter transfer method only being applicable to discrete-state-space domains (due to them using count-based exploration). \citet{parisotto16} distill policies from training tasks into a single student, which is then used to initialize the testing policy.

\section{Background}
\label{sec:background}

We model the pick and place robotics tasks in this paper as Markov Decision Processes (MDPs, \cite{bellman57}) $\mathcal{M} = \la S, A, P, R, \rho_0, \gamma \ra$. $S$ and $A$ represent the sets of state and actions, $P: S{\times}A \rightarrow Pr(S)$ is a transition function that returns a probability mass/density over states and the reward function $R: S{\times}A \rightarrow \mathbb{R}$ maps state-action pairs to their expected rewards. We consider MDPs an initial state distribution $\rho_0$ and a discount factor $\gamma$. 

A policy $\pi: S{\times}A \rightarrow [0, 1]$ captures the decision making process of an agent as the probability distribution over actions for each state. Each policy has an associated state-action value function $Q_{\pi}(s, a) = R(s, a) + \gamma \mathbb{E}_{s' \sim P, a' \sim \pi} \left[Q_{\pi}(s', a') \right]$, the discounted return when executing action $a$ in state $s$ and following policy $\pi$ thereafter.

In this paper, we consider MDPs with the following properties:

\begin{itemize}
    \item States are represented as images,
    \item the action space is large, usually one action per state pixel,
    \item the time horizon is short, around 10 time steps, and
    \item rewards are sparse.
\end{itemize}

Learning an optimal policy for this class of MDPs without additional information is extremely difficult because there is only a handful of optimal actions in each state. Hence, the probability of getting a reward for a sequence of random actions is minuscule.

\begin{algorithm}[t!]

\caption{Action prior learning}\label{alg:learnap}

\begin{flushleft}
    \hspace*{\algorithmicindent} \textbf{Input:} Set of training tasks $T = \{ T_1, T_2, ..., T_N \}$. \\
    \hspace*{\algorithmicindent} \textbf{Output:} Action prior network $f_\text{AP}$. \\
\end{flushleft}

\begin{algorithmic}[1]

    \Procedure{LearnAP}{}
    
        \For {$T_i$ in $T$}
            \State Train an expert policy $\pi_i$ for task $T_i$.
            \State \parbox[t]{180pt}{%
            Collect $K$ transitions by rolling out $\pi_i$. Store visited states in $D_i$. \strut}
        \EndFor
        \State Concatenate datasets $\{ D_1, D_2, .., D_N \}$ into $D$.
        \State Train task classifier $f_\text{C}: S \rightarrow \Delta^{N - 1}$ on $D$ (Section \ref{sec:methods:task_classifier}).
        \State Collect optimal action sets for $\pi_1, \dots, \pi_N$ on $D$.
        \State \parbox[t]{200pt}{%
        Merge optimal action sets using $f_\text{C}$ and add the union set to $D$ (Section \ref{sec:methods:masks}). \strut}
        \State Train action prior network $f_\text{AP}$ on $D$ (Section \ref{sec:methods:prior}).
        \State Return $f_\text{AP}$.
    
    \EndProcedure
    
\end{algorithmic}

\end{algorithm}

\section{Problem statement}
\label{sec:problem}

Let $T_\text{train} = \{ T_1, T_2, ..., T_N \}$ be a set of training tasks expressed as MDPs. A task $T_i = \la S, A, P_i, R_i, \rho_0, \gamma \ra$ shares its definition with all other tasks except for its reward and transition function. For example, a task of building a tower from blocks of height two and a task of building a tower of height three clearly have a different reward function. Even though the dynamics of picking and placing objects are the same for both tasks, the former task terminates when a tower of two is built, whereas the latter does not. Therefore, there are small variations in the transition function between the tasks related to terminal states.

We assume we have access to an expert policy $\pi_i$ for each training task $i$ together with a dataset of on-policy transitions $D_i$. Given a testing task $T_{N+1}$ (different in its transition and reward dynamics from training tasks), our goal is to learn the best possible policy. We formalize this as summarizing experience from previous tasks $\left( D_i, \pi_i \right)_{i=1}^N$ in some function (such as an action prior) parameterized by $\phi$. The parameters are then used in some training process $\pi(\phi)$ resulting in a policy for the testing task. We then indirectly maximize the success of the testing policy by manipulating $\phi$.
\begin{align}
    \argmax_{\phi} \mathbb{E}_{\pi(\phi), \rho_{0}, P_{N+1}} \left[ \sum_{t=0}^\infty \gamma^t R_{N+1}(s_t, a_t) \right].
\end{align}

\begin{algorithm}[t!]

\caption{Action prior exploration}\label{alg:exploreap}

\begin{flushleft}
    \hspace*{\algorithmicindent} \parbox[t]{220pt}{%
    \textbf{Input:} Reinforcement learning agent $f_\text{\:RL}$, action prior network $f_\text{AP}$, action prior probability threshold $\sigma$, task $T$. \strut} \\
\end{flushleft}

\begin{algorithmic}[1]

    \Procedure{ExploreAP}{}
    
        \While {Stopping condition not reached}
            \State Get environment state $s$.
            \If {Explore}
                \State $\bar{A}^* \leftarrow \big\{a ~\big\vert~ a \in A \wedge f_\text{AP}(s,a) > \sigma \big\}$.
                \State Randomly sample $a \sim \text{Uniform}(\bar{A}^*)$.
            \Else
                \State \parbox[t]{180pt}{%
                Choose $a$ according to $f_\text{\:RL}$ (e.g. $a$ with maximum Q-value in DQN). \strut}
            \EndIf
            \State Execute action $a$ in the environment.
            \State Observe reward $r$ and next state $s'$.
            \State Add tuple $(s, a, r, s')$ into the replay buffer.
            \State Perform a learning step of $f_\text{\:RL}$.
        \EndWhile
    
    \EndProcedure
    
\end{algorithmic}

\end{algorithm}

\section{Action Priors}
\label{sec:methods}

Given a set of expert policies $\pi_1, \dots, \pi_N$ for training tasks $T_i \in T_\text{train}$, we define the action prior to be a policy
\[
\pi_\text{AP}(s,a) = \eta \max_{i \in [1 \dots N]} \pi_i(s,a),
\]
where $\eta$ is normalizes $\pi_\text{AP}$. For example, if $\pi_1, \dots, \pi_N$ are deterministic, then $\pi_\text{AP}$ assigns equal probability to each action $\pi_1(s) \dots \pi_N(s)$.

Algorithm \ref{alg:learnap} outlines the procedure we use to train the action prior. First, we train expert policies for the $N$ training tasks. Action priors are invariant to the method used to train them (Section \ref{ap:sec:expert_policies}). Then in Step 4, for each task $i$ and policy $\pi_i$, we obtain a sample of on-policy states by rolling out $\pi_i$. 

Next, in Step 7, we train a classifier $f_{C}$ that predicts the task that is most likely to have caused the agent to visit a state. This is important because the policies $\pi_i$ are not all valid over the entire state space. For example, a policy trained to assemble the structure in Figure \ref{fig:block_picking_setup}a (tower from two cubes and a small roof) has never seen the state shown in Figure \ref{fig:block_picking_setup}b (a house built from two blocks, a brick and a large roof). To determine the set of policies applicable in a given state, we train a task classifier $f_{C}: S \rightarrow \Delta^{N-1}$ (Section \ref{sec:methods:task_classifier}), which predicts the tasks in which a state is most frequently encountered. This allows the action prior to ignore policies for tasks that are not relevant to a particular state.

Then, in Step 8 of Algorithm~\ref{alg:learnap}, after training the policies $\pi_i$ and the task classifier $f_{C}$, we collect the training dataset for the action prior (Section \ref{sec:methods:masks}). The dataset contains an equal number of states for each task, which are obtained by rolling out the learned policies $\pi_i$. For each state, we compute a binary mask that represents the union of actions that are optimal (in any task) given a set of policies. The set of applicable policies is predicted by the task classifier. Step 10 of Algorithm~\ref{alg:learnap} trains the action prior network $f_\text{AP}: S{\times}A \rightarrow [0, 1]$ to predict the probability of an action being optimal for any task in a given state (Section \ref{sec:methods:prior}).
Finally, we create an action prior policy $\pi_\text{AP}(s, a)$ based on the action prior network $f_\text{AP}$. By thresholding the probabilities predicted by $f_\text{AP}$, we get a set of proposed actions $A^*$(s) for state $s$. We set $\pi_\text{AP}(s, a)$ to be a uniform distribution over $A^*(s)$ with actions outside of the optimal set being assigned zero probability.


\subsection{Learning the Task Classifier}
\label{sec:methods:task_classifier}

The task classifier $f_{C}: S \rightarrow \Delta^{N-1}$ determines which of the expert policies are relevant in the context of a particular state. To train this classifier, we use a dataset of states and categorical labels $\{ (s_i, y_i) \}_{i=1}^M$. We construct this dataset by generating policy rollouts for each task. If a state $s$ was visited during a rollout of a policy for a task $y \in \{1, \dots, N\}$ we include the pair $(s,y)$ in the dataset. Note that this results in a dataset where class labels for each state are not unique, as each state could have been encountered during rollouts for multiple tasks (e.g.~the initial state with all objects placed on the ground appears in all tasks). We can interpret the data as samples $s_i, y_i \sim p(s, y)$ from a distribution $p(s,y) = p(s \,|\, y) p(y)$ in which $p(y)$ is a uniform prior over tasks (i.e.~the training dataset is balanced), and $p(s \,|\, y)$ is the fraction of rollouts for each task in which state $s$ appears. The classifier now approximates the conditional distribution $p(y \,|\, s) \propto p(s, y) \propto p(s \,|\, y)$.


We implement $f_{C}$ as a neural network $\text{NN}(s)$ that predicts logits for all classes, which we normalize using a softmax function 
\begin{align}
    p(y \mid s) \simeq f_{C}(s) = \text{softmax}(\text{NN}(s)),
\end{align}
and train the classifier using a cross-entropy loss
\begin{align}
    L_C = - \frac{1}{M} \sum_{i=1}^M \sum_{j=1}^{|T|} \mathbb{I}[y_i = j] \log f_{C}(s)_j.
\end{align}
To determine if policy for task $j$ is applicable in state $s$, we check if the predicted probability $f_{C}(s)_j$ is above some threshold $\delta$. 

Since neural classifiers in general do not have well-calibrated probabilities, our approximation of $p(y \mid s)$ can be overconfident \cite{guo17}. That said, we find that this classification strategy works sufficiently well for our purposes in practice.

\subsection{Approximating Optimal Action Sets}
\label{sec:methods:masks}


For an ideal action prior, we wold like to determine the set of actions $A^*(s)$ that are optimal for \emph{any} task in state $s$. Given an expert policy $\pi_i$ for task $i$ and corresponding value function $Q_{\pi_i}$, we define the optimal action set for state $s$ as
\begin{align}
    A^*(s) =
    \bigcup_{i=1}^N
    \left\{ 
        a 
        \hspace{0.5em}\middle\vert\hspace{0.5em}
        Q_{\pi_i}(s, a)
        =
        \max_{a' \in A} \: 
        Q_{\pi_i}(s, a')
    \right\}.
\end{align}

We expect the cardinality of $A^*(s)$ to be high in our domain, as there are many equivalent ways of picking and placing objects. Since our learned expert policies tend to be noisy, it is however non-trivial to determine the set of optimal actions without carefully setting a threshold for each trained model.

Instead, we restrict the optimal actions set to one action per task with the optimal action for the $i$th task denoted by $a^*_i = \argmax_a \pi_i(s, a)$ with ties broken randomly. These action form an approximate optimal actions set
\begin{align}
    \widetilde{A}^*(s) = \{ a^*_1, a^*_2, ..., a^*_N \}.
\end{align}

$\widetilde{A}^*(s)$, which we can compute, is a subset of the ground-truth set $A^*(s)$. We deal with the problem of increasing the number of proposed optimal actions in Section \ref{sec:methods:prior}. Furthermore, we restrict the optimal actions set to only the task determined to be applicable by the task classifier (Section \ref{sec:methods:task_classifier}).

\subsection{Learning the Action Prior}
\label{sec:methods:prior}

We use the approximate optimal action sets $\widetilde{A}^*(s)$ to learn an action prior $f_\text{AP}: S{\times}A \rightarrow [0, 1]$, which takes the form of a multi-task classifier that returns binary predictions for all actions $A$ given a state $s \in S$. We train this classifier using pairs of states and optimal actions masks $\{ (s_i, m_i) \}_{i=1}^L$. We collect training states by rolling out the learned policies for each task for a fixed number of time steps. For each state, we then compute the optimal mask, which for each action $a$ contains a bit that indicates whether this action is part of the approximate optimal action set
\begin{align}
    m_{i,a} = \mathbb{I} \left[a \in \widetilde{A}^*(s_i) \right].
\end{align}
We train this multi-task classifier using a standard logistic loss
\begin{align}
    \nonumber L_\text{AP} = - \frac{1}{L} \sum_{i=1}^L \sum_{j=1}^{|A|} &m_{ij} \log f_\text{AP}(s_i, a_j) \: + \\
    &(1 - m_{ij}) \log \left(1 - f_\text{AP}(s_i, a_j) \right).
\end{align}

As discussed above, the training masks contain only a subset of the union of optimal action sets for all tasks. We can view this problem from the perspective of precision-recall trade-off. A model that achieves a low $L_\text{AP}$ value will have high precision (i.e., it will avoid false-positive optimal actions), but it might have low recall because not every optimal action is represented in the training data. 

This trade-off can be controlled by moving the decision boundary $\mathbb{I}[f_\text{AP}(s, a) \geq \sigma]$ that determines if an action is deemed optimal. Experimentally, we found that setting $\sigma$ to a low value (i.e., increasing recall and decreasing precision) results in a large increase in the success rate of the action prior exploration policy.

\subsection{Action prior exploration}
\label{sec:methods:exploration}

Algorithm \ref{alg:exploreap} summarizes how we use the action prior for exploration when training a reinforcement learning agent on a new task. We follow the standard recipe of alternating between selecting random actions (exploration) and action according to the policy being learned (exploitation). The most common and one of the simplest approaches to controlling this trade-off is an $\epsilon$-greedy exploration policy with the value of $\epsilon$ decaying over time (e.g., this approach was used in the original deep Q-network paper \cite{mnih15}). 

We use an $\epsilon$-greedy strategy in which, during exploration, we select actions uniformly at random from the set actions for which the action prior exceeds a threshold $\sigma$
\begin{equation}
    \bar{A}^*(s) = \big\{a ~\big\vert~ a \in A \wedge f_\text{AP}(s,a) > \sigma \big\}.
\end{equation}
This results in a more focused exploration compared to selecting actions uniformly at random from the full set $A$. 

It is possible that a new task will require the agent to select an action that was not optimal in any previous tasks. In this context, it could be beneficial to occasionally select a completely random action during the exploration step. But, we find it to decrease the success rate of the action prior exploration policy. We hypothesize that by setting $\sigma$ to a low value (around 0.1) the action prior generalizes to actions that were not necessarily optimal for any previous task but are nevertheless plausible.

\begin{figure}
    \centering

    \begin{subfigure}[t]{1\columnwidth}
        \centering
        \includegraphics[width=1\textwidth]{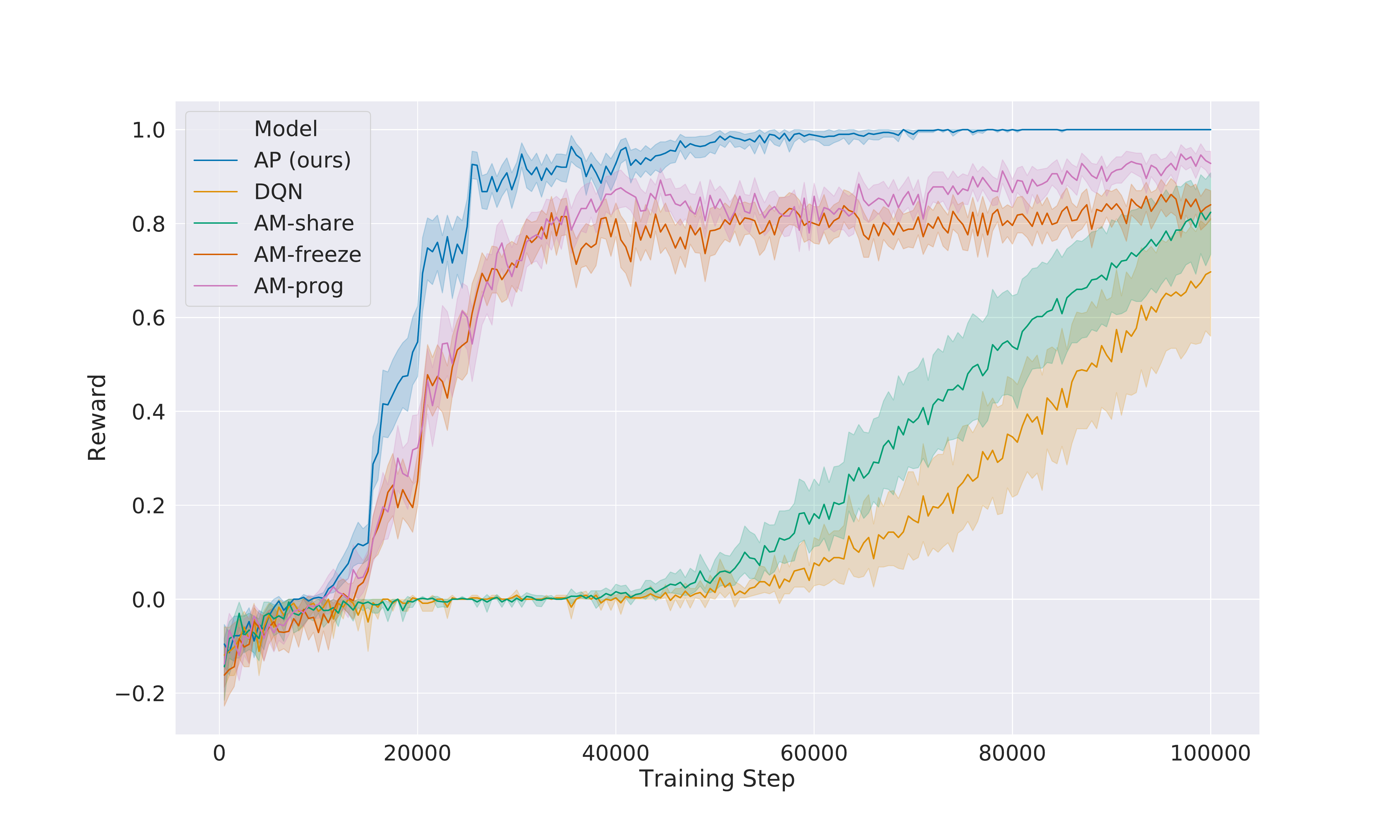}
        \caption{Pick up four fruits in any order.}
    \end{subfigure}
    \begin{subfigure}[t]{1\columnwidth}
        \centering
        \includegraphics[width=1\textwidth]{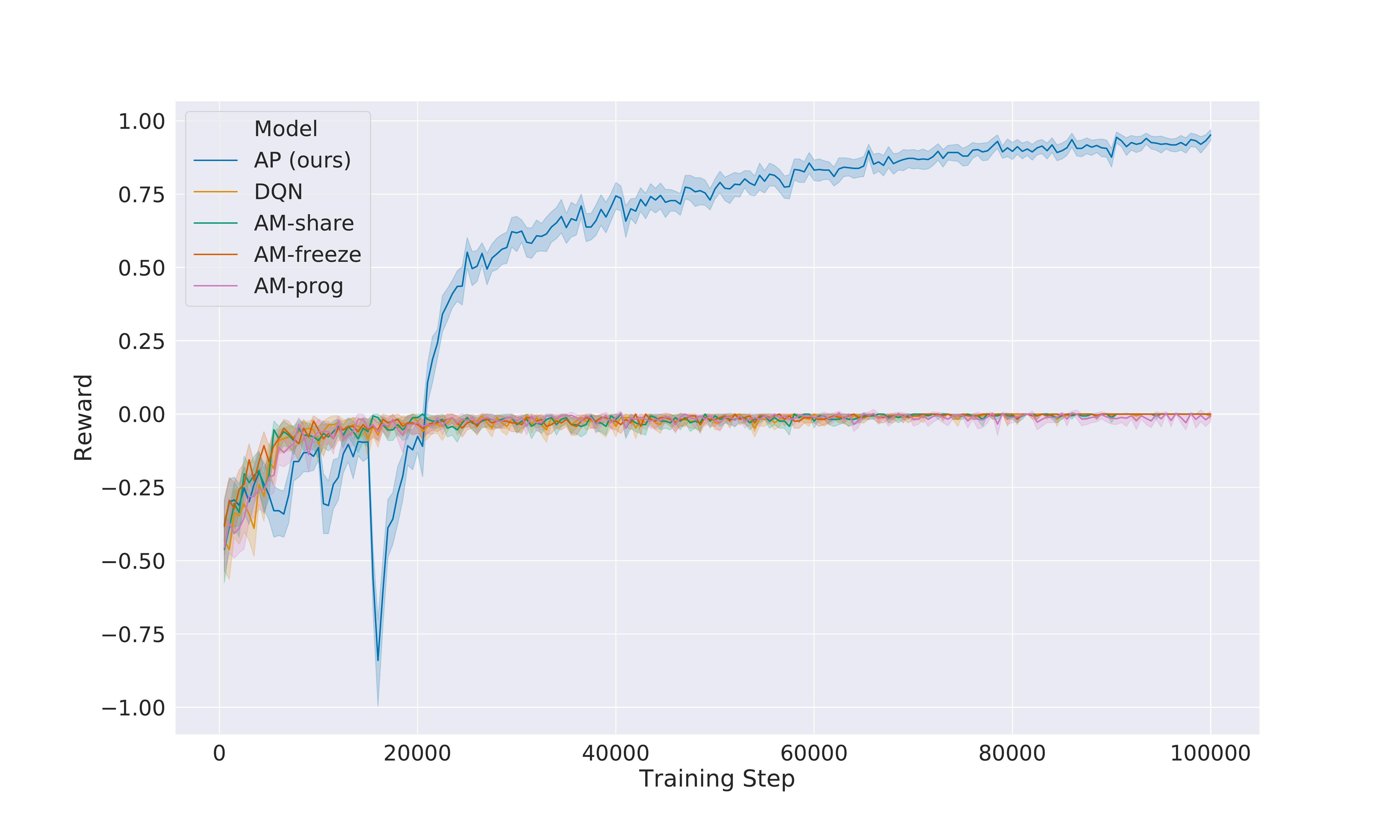}
        \caption{Pick up four fruits in a specific order.}
    \end{subfigure}

    \caption{Transfer learning results in Fruits World. There are five distinct fruits in the environment and the agent should pick up between one and four of them. Picking can be done either (a) in any order or (b) in a particular order. We use all possible tasks for fruit combinations (30 tasks) and sample 20 tasks for fruit sequences. The results were obtained by leave-one-out cross-validation, where we learn an action prior over $N-1$ tasks and perform transfer learning on the $N$th task. We plot the learning curves for the hardest tasks here, see Table \ref{tab:fruits_seq} in the Appendix for all results. We report means and its 95\% confidence intervals over 10 runs.}
    \label{fig:fruits_learning_curves}
\end{figure}

\section{Experiments}
\label{sec:experiments}

We demonstrate the effectiveness of action priors in two domains: a proof-of-concept Fruits World (Section \ref{sec:experiments:fruits}) and a block stacking robotic manipulation experiment in the PyBullet physics simulator (Section \ref{sec:experiments:blocks}). Policies learned in PyBullet can be deployed on a real-world UR5 robotic arm (Section \ref{sec:experiments:robots}). The key question we aim to answer is if action priors enable a deep Q-network (DQN) to learn tasks that were previously out of reach.

\subsection{Domains}
\label{sec:experiments:domains}

\begin{table*}[t!]
    \centering
    \begin{tabular}{ccccccccc}
        \toprule
        Method & \multicolumn{8}{c}{Final success rate on task} \\
         & 1b1r & 2b1r & 2b2r & 1l1r & 1l2r & 1b1b1r & 2b1b1r & 2b2b1r \\
         & \includegraphics[width=4em]{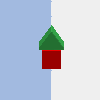} 
         & \includegraphics[width=4em]{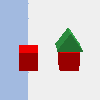} 
         & \includegraphics[width=4em]{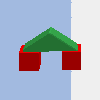}
         & \includegraphics[width=4em]{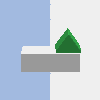}
         & \includegraphics[width=4em]{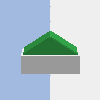}
         & \includegraphics[width=4em]{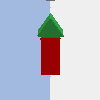}
         & \includegraphics[width=4em]{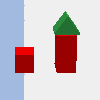}
         & \includegraphics[width=4em]{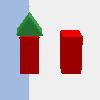} \\
        \midrule
        DQN RS & 100\% & 0\% & 0\% & 100\% & 100\% & 0\% & 0\% & 0\% \\
        DQN RS, WS & 98\% & 0\% & 0\% & 99\% & 100\% & 0\% & 0\% & 0\% \\
        DQN HS & 97\% & 0\% & 0\% & 100\% & 100\% & 0\% & 0\% & 0\% \\
        DQN HS, WS & 97\% & 0\% & 2\% & 100\% & 99\% & 3\% & 0\% & 0\% \\
        \textbf{DQN AP} & 100\% & 100\% & 98\% & 99\% & 100\% & 100\% & 93\% & 0\% \\
        \textbf{DQN AP, WS} & 100\% & 96\% & 97\% & 99\% & 99\% & 99\% & 92\% & 0\% \\
        \midrule
         & 2b2b2r & 2b1l1r & 2b1l2r & 1l1b1r & 1l2b1r & 1l2b2r & 1l1l1r & 1l1l2r \\
         & \includegraphics[width=4em]{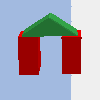} 
         & \includegraphics[width=4em]{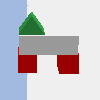} 
         & \includegraphics[width=4em]{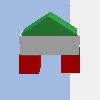}
         & \includegraphics[width=4em]{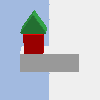}
         & \includegraphics[width=4em]{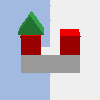}
         & \includegraphics[width=4em]{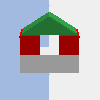}
         & \includegraphics[width=4em]{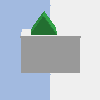}
         & \includegraphics[width=4em]{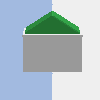} \\
        \midrule
        DQN RS & 0\% & 0\% & 0\% & 0\% & 0\% & 0\% & 0\% & 0\% \\
        DQN RS, WS & 0\% & 0\% & 0\% & 0\% & 0\% & 0\% & 0\% & 0\% \\
        DQN HS & 0\% & 0\% & 0\% & 10\% & 0\% & 0\% & 5\% & 92\% \\
        DQN HS, WS & 0\% & 0\% & 0\% & 3\% & 0\% & 0\% & 3\% & 96\% \\
        \textbf{DQN AP} & 95\% & 96\% & 0\% & 100\% & 99\% & 90\% & 93\% & 97\% \\
        \textbf{DQN AP, WS} & 0\% & 97\% & 0\% & 100\% & 100\% & 98\% & 100\% & 99\% \\
        \bottomrule
    \end{tabular}
    \vspace{0.5em}
    \caption{Transfer experiments in the block stacking domain. We consider "DQN AP, WS" the main contribution of this paper. Each column reports the final success rate averaged over 100 episodes after training. The baselines and ablations of our method we consider are random action selection (RS), heuristic action selection (HS), weight sharing (WS) and action prior (AP).}
    \label{tab:blocks_transfer}
\end{table*}

\textbf{Fruits World} is a grid-world-like domain. The world is a $5{\times}5$ grid with five distinct fruits placed in random positions (the positions change at the start of each episode). There are 25 actions, one for each position. The agent must pick a subset of fruits either in a particular order (sequences) or in any order (combinations); each subset constitutes a different task. When the agent thinks it finished the task, it must execute the 25th action to get a reward and reset the environment. A reward of 1 is given for successfully picking up the target fruits. Due to its combinatorial nature, this environment is 
\begin{wrapfigure}[14]{r}{0.2\textwidth}
 \vspace{-0.25cm}
  \begin{center}
    \includegraphics[width=0.2\textwidth]{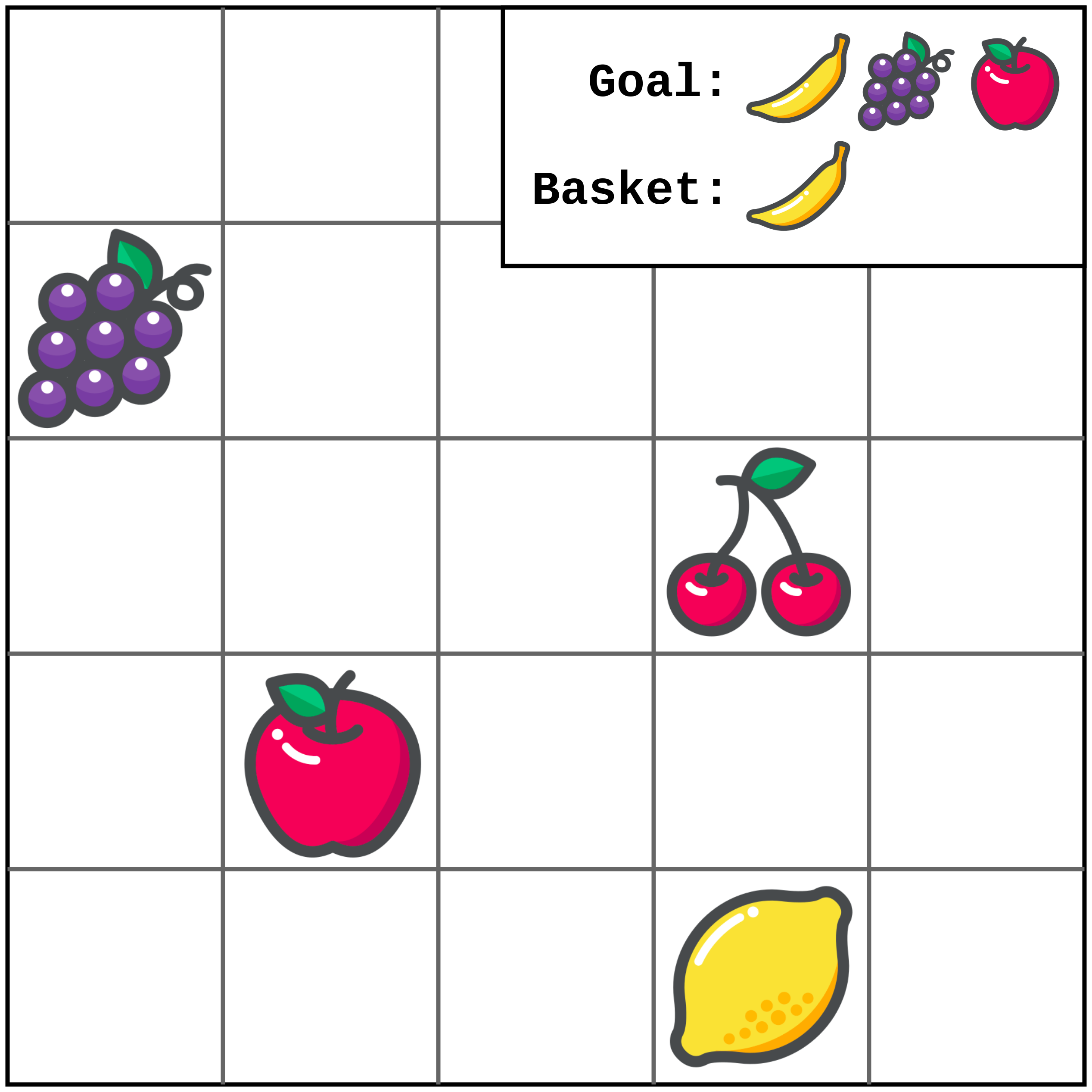}
    \caption{Sequential fruit picking task.}
    \label{fig:fruits_viz}
  \end{center}
\end{wrapfigure}
surprisingly difficult to solve with model-free deep reinforcement learning. Hence, we also give the agent a reward of -0.1 for picking up the wrong fruit and only allow it to put the right fruits in the basket.

The state is a $5{\times}5{\times}5$ tensor with the first two dimensions corresponding to the grid. If fruit is present in a grid cell, its ID is one-hot encoded in the third dimension. Otherwise, the values in the grid cell are all zero. In combinations, a sixth channel is added to the observation: the positions corresponding to fruits the agent has added to its basket are set to one. In sequences, the agent is given an additional sequence of one-hot encoded IDs of fruits it has picked.

We use a deep Q-network as a model-free agent. All neural networks are a multilayer perceptron with two hidden layers (see \ref{ap:sec:exp_details:fruits} in the Appendix).

\textbf{PyBullet block stacking} is a set of simulated tasks involving stacking blocks of various shapes and sizes with a robotic arm (Figure \ref{fig:block_picking_setup}). The robot observes the workspace with a depth camera from above (Figure \ref{fig:block_picking_setup} (b)), receiving a $90{\times}90$ image with pixel values corresponding to heights. It can execute a top-down pick or place action at a specified coordinate with fixed hand rotation. Before executing a pick action, the robot will take a $24{\times}24$ picture centered at the coordinates, where it is executing the pick action. If it successfully picks up an object, it uses the in-hand image to decide where to place it. 

We discretize the action space as a $90{\times}90$ grid, each cell corresponding to one pixel of the observation. We instantiate 16 different block stacking tasks: each one builds a structure of a width of one or two small blocks and a height of two or three blocks; each structure has a roof on top. Figure \ref{fig:block_picking_setup} (c) and (d) show goal states of building a tower from two small blocks and a small roof and of building a structure from two small blocks followed by one long block and a large roof respectively. Each task is represented as a string (e.g. a small roof on top of a small block is "1b1r") and we use a context-free grammar to generate all possible tasks with particular parameters (Appendix Section \ref{ap:sec:grammar}).

Both the $90{\times}90$ observation and the $24{\times}24$ in-hand image are encoded using a modified version of U-Net \cite{ronneberger15,wang2020policy}. We chose this model because it can produce detailed segmentation maps. The task our DQN, which we use for model-free learning, and action prior models solve is comparable to image segmentation: we make a prediction for each pixel of the input image. In the DQN case, the U-Net predicts a single state-action value for each pixel of the input image. In the action prior case, the model produces a logit for each pixel of the observation. The exact architecture of our U-Net is depicted in Figure \ref{ap:fig:unet} in the appendix.

\begin{figure}
    \centering

    \begin{subfigure}[t]{0.48\columnwidth}
        \centering
        \includegraphics[width=0.828\textwidth]{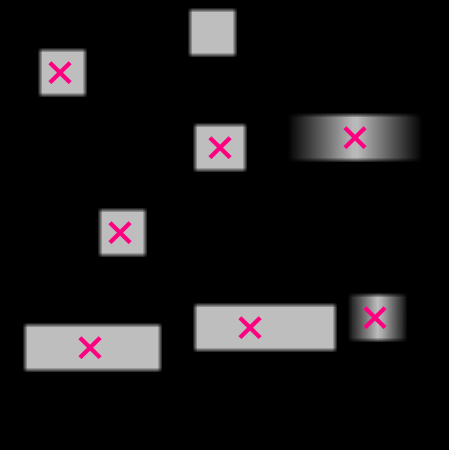}
        \caption{$\sigma=0.1$, empty hand}
    \end{subfigure}
    \begin{subfigure}[t]{0.48\columnwidth}
        \centering
        \includegraphics[width=0.828\textwidth]{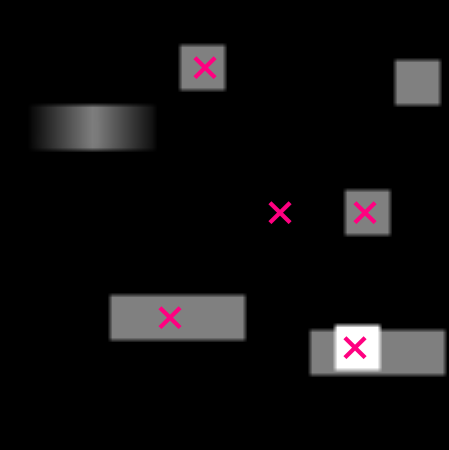}
        \caption{$\sigma=0.1$, cube in hand}
    \end{subfigure}
    \begin{subfigure}[t]{0.48\columnwidth}
        \centering
        \includegraphics[width=0.828\textwidth]{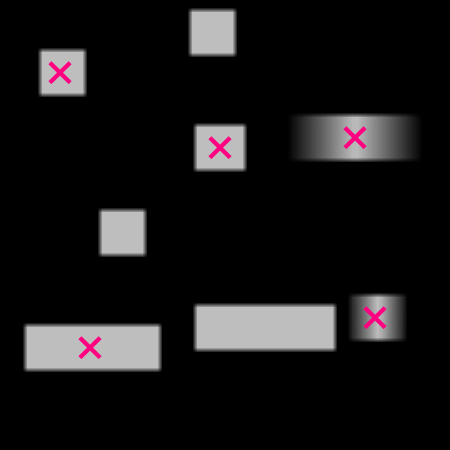}
        \caption{$\sigma=0.5$, empty hand}
    \end{subfigure}
    \begin{subfigure}[t]{0.48\columnwidth}
        \centering
        \includegraphics[width=0.828\textwidth]{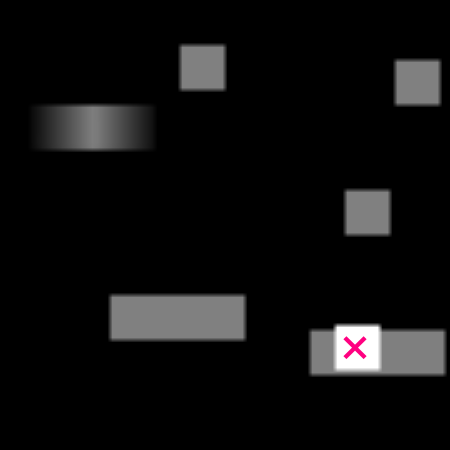}
        \caption{$\sigma=0.5$, cube in hand}
    \end{subfigure}

    \caption{Visualizing actions proposed by an action prior (red crosses) with the probability threshold $\sigma$ set to 0.1 (top row) and 0.5 (bottom row). In the first column, the robot is supposed to pick up an object, in the second column, it is holding a cube in its hand and wants to place it. The images shown are pictures from a depth camera mounted at the top of the workspace in simulation.}
    \label{fig:actions}
\end{figure}

\subsection{Fruits World experiments}
\label{sec:experiments:fruits}

We sample 20 and 30 tasks for fruits sequences and combinations tasks respectively (Section \ref{sec:experiments:domains}). Action priors are tested in a leave-one-out transfer experiment: an action prior is trained on 19 and 29 tasks respectively; it is then used for exploration on the held-out task. This process is repeated for each task. Expert policies for the training tasks use imitation learning (Section \ref{ap:sec:expert_policies}).

Our comparison in  Figure \ref{fig:fruits_learning_curves} involves action priors (AP), a randomly initialized deep Q-network (DQN) and three baselines based on Actor-Mimic \cite{parisotto16} (AP-share, AP-freeze and AP-prog). Actor-Mimic first trains a student to mimic the policies of all experts. The student weights are then used to initialize the agent when learning the testing task. We followed the original implementation\footnote{URL: \url{https://github.com/eparisotto/ActorMimic}, visited 21/02/08.} except for us having a student with $N$ heads, one for each training task. This deviation from the original paper is due to the state of our environment not containing any information about which task the agent is solving. In contrast, \citet{parisotto16} test their method on several Atari games, each with a distinct state space. 

In all cases, we remove the $N$ student heads and transfer the weights of the hidden layers. In AM-share, all weights are trainable; AM-freeze only learns the final fully-connected layer. AM-prog is based on \citet{rusu16}, where each hidden layer of the agent network receives both features from the previous agent layer and from the previous student layer. Hence, there are two networks, student and agent, but only the agent is trained. We do not use an adaptation layer suggested by \cite{rusu16} because we only have one student network.

Action priors significantly outperform all baseline in fruits combinations (Figure \ref{fig:fruits_learning_curves}a) and are the only method to solve any fruits sequences task (Figure \ref{fig:fruits_learning_curves}b). In the former, AM-prog and AM-freeze have similar performance, whereas random initialization and AM-share train much slower. We hypothesize that AM-share overwrites the student weights at the start of training, causing it to perform similarly to random initialization. The negative rewards at the start of training received by action priors in Figure \ref{fig:fruits_learning_curves}b can be explained by AP picking up the wrong fruits (small negative reward), and the other methods failing to pick up any fruit (zero reward).

\subsection{PyBullet block stacking experiments}
\label{sec:experiments:blocks}

\begin{figure}[t!]
    \centering

    \begin{subfigure}[t]{1.0\columnwidth}
        \centering
        \includegraphics[width=0.7\textwidth]{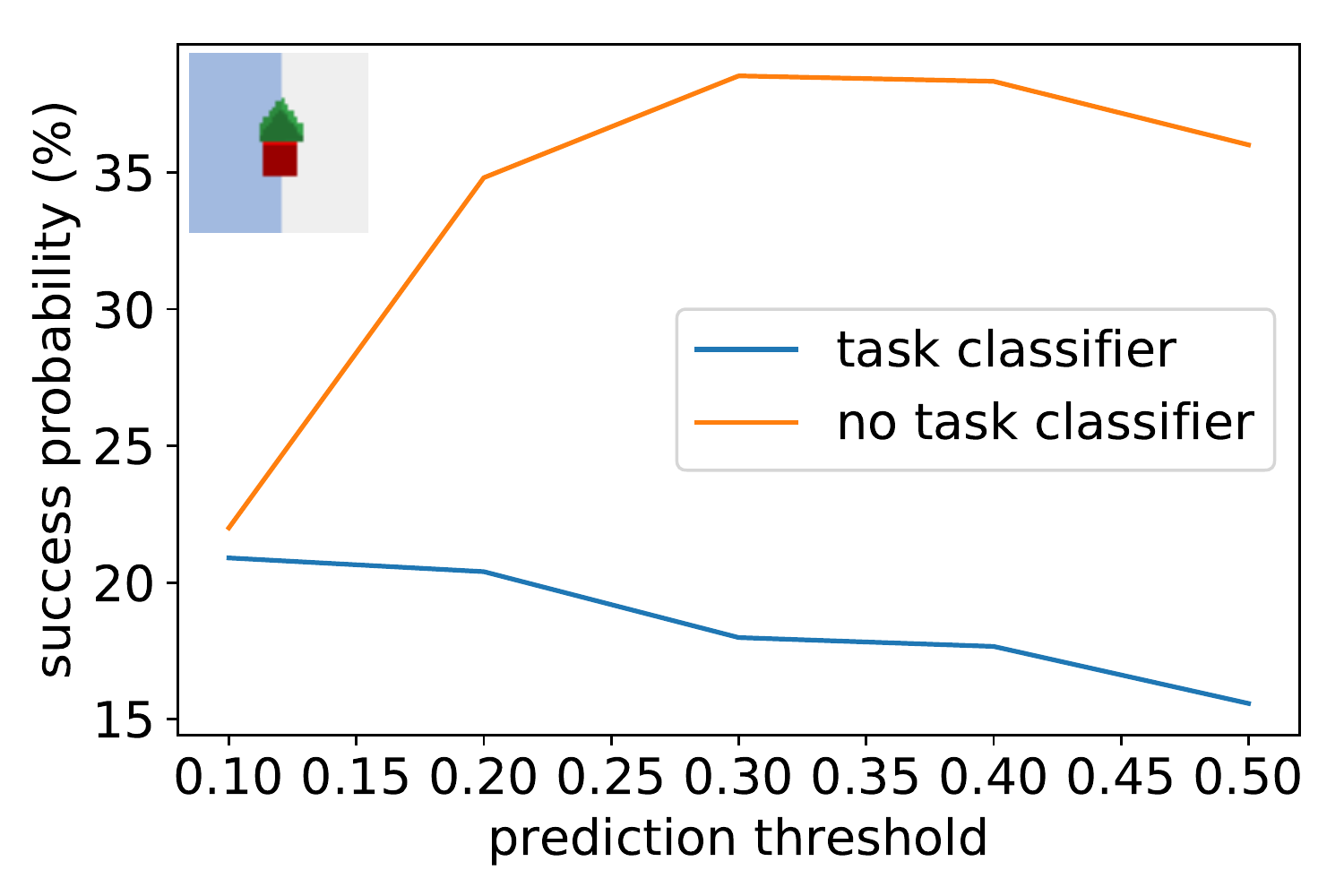}
    \end{subfigure}
    \begin{subfigure}[t]{1.0\columnwidth}
        \centering
        \includegraphics[width=0.7\textwidth]{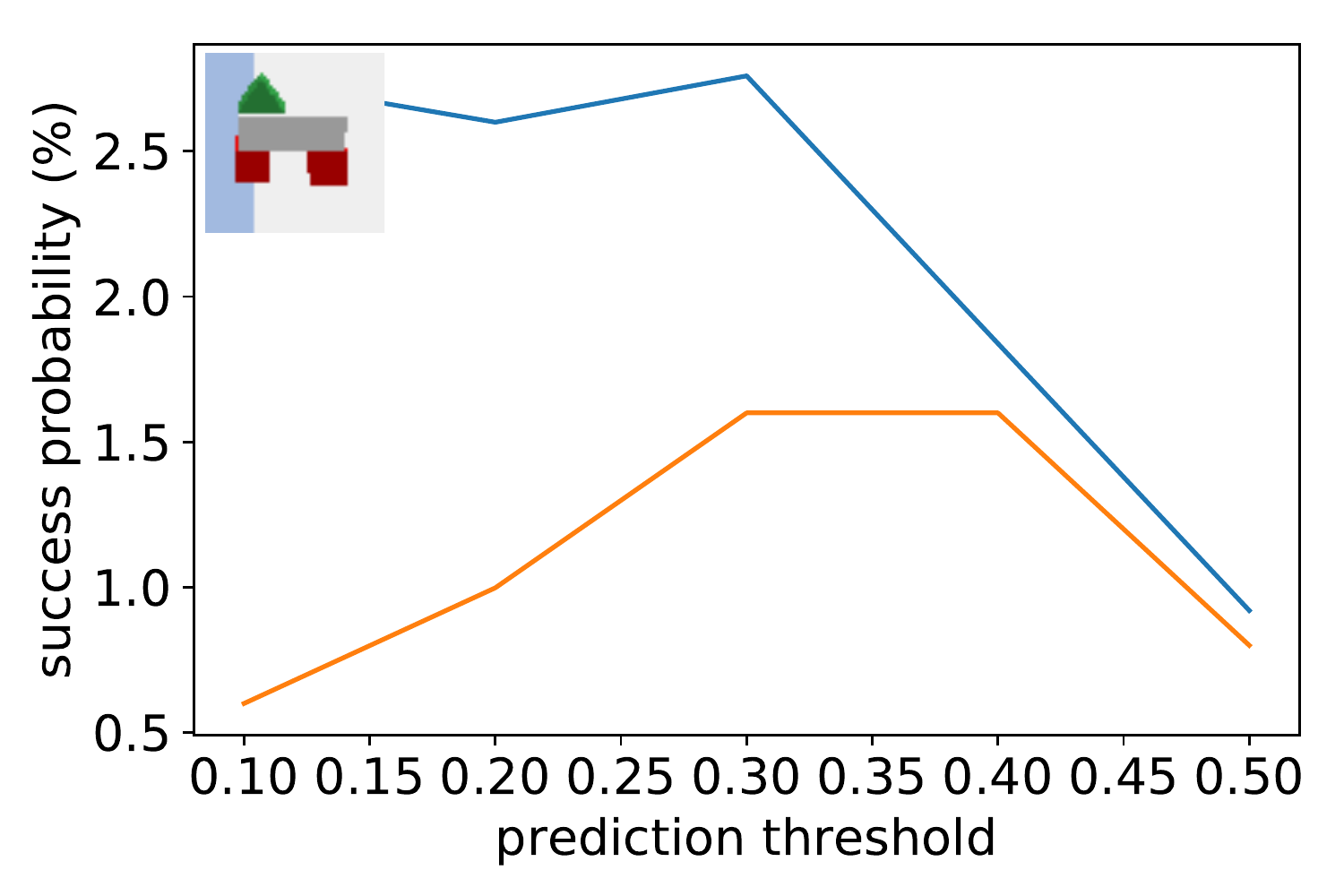}
    \end{subfigure}

    \caption{Success rates for "1b1r" and "2b1l1r" using exploration with an action prior as a function of $\sigma$. We compare action priors trained with and without task classifier.}
    \label{fig:block_exp}
\end{figure}

We perform two experiments in this domain: one focused on model-free learning with action priors and the other solely on exploration using action priors.  The first experiment matches the setup in Section \ref{sec:experiments:fruits}: we use 15 out of the 16 block stacking tasks to learn an action prior, which we subsequently test on the 16th task (Table \ref{tab:blocks_transfer}). An imitation learning method called SDQfD facilitates the expert policies for the training tasks (Section \ref{ap:sec:expert_policies}). The second experiment evaluates the success rate of action prior policies on all tasks without additional training (Figure \ref{fig:block_exp} shows two example tasks, Figure \ref{ap:fig:exp_all} contains the results for all tasks). We measure success rate as a function of the action prior probability threshold $\sigma$ and the presence or absence of a task classifier.

\textbf{Experiment \#1} compares our method with two baselines. \textbf{DQN AP} is a deep Q-network with action prior exploration. We test random and heuristic action selection. As our observations are depth images taken from the top of the workspace, the heuristic only allows selecting actions that act on parts of the observation with non-zero height. Therefore, it forces the agent to interact with objects. But, not all interactions necessarily lead to desirable outcomes--the agent often ends up pushing objects out of the workspace. We call the two methods \textbf{DQN RS} and \textbf{DQN HS} respectively.

As a second mechanism for speeding up learning, we employ weight sharing in the unseen task. Because an action prior summarizes the information contained in experts for the 15 training tasks, we initialize the DQN for the 16th testing task with the action prior weights. To mitigate catastrophic forgetting \cite{french99}, we include a $L^2$ penalty between the original action prior weights and the current DQN weight; it is added to the DQN loss function with a weight of $\omega_{WS}$. We test weight sharing both with random and heuristic baselines (\textbf{DQN RS, WS} and \textbf{DQN HS, WS}) and action priors (\textbf{DQN AP, WS}).

A DQN with an action prior can solve all tasks except for "2b2b1r" and "2b1l2r" (Table \ref{tab:blocks_transfer}). As shown in Figure \ref{ap:fig:exp_all}, the success rates of action priors on these tasks without training is between $1$ and $4\%$. Hence, a more targeted action prior might be needed.

The DQN sharing weights with the action prior does not lead to noticeable benefits. In fact, it fails to learn "2b2b2r" unlike the non-weight-sharing variant. Since the purpose of weight sharing is to improve training speed of the network, it only provides minor benefits due to exploration being the main challenge in our domains.

As we expected, both the heuristic and random action selection baselines are unable to solve most of the tasks. Random actions succeed on the three simplest tasks ("1b1r", "2b1r" and "2b2r"), and heuristic selection increases the number to four (plus "1l1l2r"). In the rest of the tasks, the success rates of the baseline exploration policies are close to zero; therefore, no learning method including weight sharing can succeed.

\textbf{Experiment \#2} evaluates two variants of action prior each again trained in leave-one-out fashion on 15 tasks and tested on the 16th task. In this experiment, we only log the success rate of the action prior policy without learning on the testing task.

Figure \ref{fig:block_exp} shows examples of action prior success rates for tasks "1b1r" and "2b1l1r". We plot the success rates as a function of the action prediction threshold $\sigma$ and the presence or absence of a task classifier. The success rates are measured by rolling out episodes in the test task (depicted in the top-left corner) with actions selected by the action prior policy $\pi_{AP}$. We show results for all tasks in Appendix Figure \ref{ap:fig:exp_all}. 

The main trends in Figure \ref{fig:block_exp} (and the Appendix Figure \ref{ap:fig:exp_all}) are that (a) more complicated tasks benefit more from the task classifier and that (b) almost all action priors that use a task classifier benefit from a decrease in the prediction probability threshold. The increase in success rates for action priors with task classifiers in more complex tasks can be explained by them needing a targeted and precise exploration policy: a task classifier ensures that the training data for the action prior does not mark actions from irrelevant tasks as optimal. On the other hand, we are unsure as to why action priors without task classifiers perform well on simple tasks. See Figure \ref{fig:actions} for example of sets of actions proposed by action priors.

\subsection{Real-world robot experiments}
\label{sec:experiments:robots}

In theory, policies trained in our PyBullet simulation can be transferred to a real-world robot, as both setups use a depth camera and a robotic arm with the same action spaces. However, in practice noise in the real-world observations often confuses the policies trained with perfect depth images.

To make our models more robust, we re-train the action prior DQNs with weight sharing (DQN AP, WS) in the transfer learning experiment in Section \ref{sec:experiments:blocks} with two modifications to the environment. First, we add Perlin noise, a type of correlated gradient noise, to the simulated depth images \cite{perlin85,james17}. The noise models blobs of lower and higher estimated depth values that are present in the real-world images. Next, we initialize all objects in the environment with small rotations compared to their default orientations. In our previous experiments, all objects had the same orientation, as the agent cannot control the rotation of the robot hand. However, the initialization of the real-world scene will inevitably be less precise.

Our setup is depicted in Figure \ref{fig:ur5} and described in details in Appendix Section \ref{ap:sec:exp_details:blocks_real}. We pick 8 of the 16 trained policies for testing. Each such policy is run for ten episodes and we report the fraction of successful episodes (Table \ref{tab:blocks_robot}). Most policies reach around $80\% - 90\%$ success rate. Two common failure modes are the robot failing to find the small roof because of noise in the depth image and the robot accidentally knocking down a structure when placing a block. The latter cannot always be attributed to a bad policy--our gripper sometimes releases objects off-center. Since the real-world gripper is significantly larger than the simulated one, it fails if objects get pushed too close to each other.

\begin{figure}[t!]
    \centering
    \includegraphics[width=0.65\columnwidth]{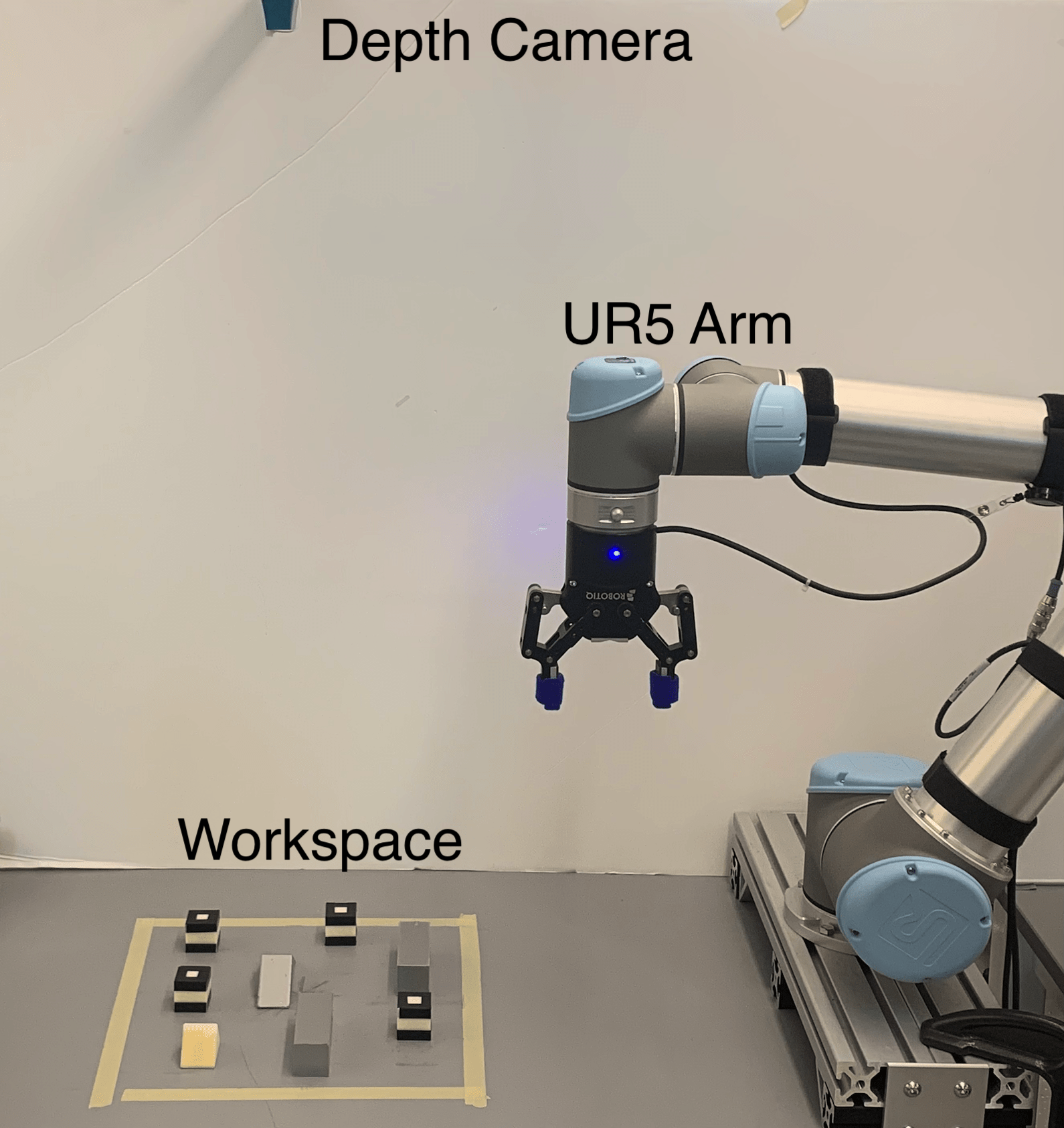}
    \caption{Our pick and place experiment with a UR5 robotic arm. The robot observes the workspace with a depth camera mounted above it, and it can choose to either pick or place an object with a top grasp and fixed orientation.}
    \label{fig:ur5}
\end{figure}

\begin{table}[]
    \centering
    \begin{tabular}{cc|cc}
        \toprule
        Task & Successes & \# Pick Failures & \# Place Failures \\
        \midrule
        1b1r & 9/10 & 1 & 0 \\
        1l1r & 8/10 & 2 & 0 \\
        1l2r & 9/10 & 1 & 0 \\
        1b1b1r & 9/10 & 1 & 0 \\
        1l1b1r & 10/10 & 0 & 0 \\
        1l2b2r & 8/10 & 1 & 1 \\
        1l1l1r & 7/10 & 1 & 2 \\
        1l1l2r & 8/10 & 0 & 2 \\
        \bottomrule
    \end{tabular}
    \vspace{0.5em}
    \caption{Real-world experiment with a UR5 robotic arm (Figure \ref{fig:ur5}). The environment was set up in the same way as the simulated workspace and we evaluated policies trained in simulation in 10 trials. Each trial had a budget of 20 time steps. We break down failures into two components: pick failures and place failures. Pick failures mean that the robot could not find the desired object or it could not pick it with sufficient precision. Place failures occur when the robot fails to place an object on the appropriate structure or it topples the structure over.}
    \label{tab:blocks_robot}
\end{table}

\section{Discussion and Conclusion}

In this work, we proposed a method for efficient exploration using action priors. Our approach to learning action priors involves solving a set of training tasks with imitation learning and summarizing the learned policies in a single action prior neural network. This network is trained on sets of optimal actions predicted by the policies with an addition of a task classifier that determines which policies are relevant in each state.

In contrast to prior work on action priors, which has predominantly considered grid-world-like environments, our method is applicable to domains with image states and thousands of actions. In addition to a proof-of-concept Fruits World domain, where it always finds near-optimal policies, we demonstrate performance on a simulated robotic block stacking task. A deep Q-network augmented with an action prior can solve 14 out of the 16 block stacking tasks within 100k episodes. Moreover, the policies trained in simulation can also be deployed on the real robotic arm with only a small drop in their success rates.

A future direction of our work is to develop action priors that are suitable to learning online during training on a new task. In this way, the action prior would progressively increase the chance of hitting the goal state while exploring. A natural extension of our manipulation task is to consider a full SE(3) action space (i.e. grasping from any angle).




\bibliographystyle{ACM-Reference-Format}
\balance
\bibliography{main}


\clearpage
\appendix
\onecolumn

\section{Stacking Grammar}
\label{ap:sec:grammar}

Instead of arbitrarily choosing tasks, we characterize the whole family of possible tasks. Here, we define a simple grammar of stacking tasks. It only captures the notion of a stack, so it cannot represent tasks such as two blocks in a row. The only notion of two blocks being near one another is if a long block or a long roof can be put on top of them.

We have the following terminals: one block (1b), two blocks next to each other (2b), one brick (1l), short roof (1r) and long roof (2r). The non-terminals are ground (G), wildcard (W), short (S) and long (L). The starting symbol is the ground.

The rules allow for stacking of long objects (two blocks next to each other or one long block) on long objects, short on short and short on long. We separate the different possible outcomes of a rule by commas.

\begin{itemize}
    \item G $\rightarrow$ 1bS, 2bW, 1lW
    \item W $\rightarrow$ S, L
    \item S $\rightarrow$ 1bS, 1b, 1r
    \item L $\rightarrow$ 1lW, 2bW, 1l, 2b, 2r
\end{itemize}

If we restrict the maximum height to three and only select structures that end with a roof, we get the 16 tasks used in our experiments.

\section{Expert Policies}
\label{ap:sec:expert_policies}

Action priors as well as our weight sharing and Actor-Mimic baselines require expert policies for \textit{training tasks}. Next, we describe the methods we used to train expert policies in Fruits World and Block Stacking, but action priors are invariant to these design decisions. In both cases, we start with a replay buffer pre-populated with expert demonstrations and learn a near-optimal policy, which we also call an expert. This might seem redundant, but, as you will see, a method that is used to generate expert demonstrations is not always able to select the optimal action in an arbitrary state. Moreover, the expert demonstrations contain optimal actions, not distributions over all actions, which are required by Actor-Mimic \cite{parisotto16}.

\textbf{Fruits World:} We use the same DQN both for the experts and for training on the \textit{testing task} (Section \ref{ap:sec:exp_details:fruits}), but the expert version is given a pre-populated replay buffer. The buffer contains 50k transitions generated by a policy that executes the optimal action with 50\% probability, and a random action otherwise. The optimal action can be easily computed given the full state of the environment. The expert is trained for 50k steps \textit{offline} on this buffer; then, it is trained \textit{online} for another 50k steps with new experiences mixed into the pre-populated buffer.

\textbf{Simulated Block Stacking:} We use an imitation learning method called SDQfD \cite{wang2020policy} to learn expert policies. Similarly to above, it requires a replay buffer pre-populated with expert demonstrations. Following \citet{wang2020policy}, we start in the goal state (i.e., with a simulator initialized so that a particular goal structure is build), pick blocks from the top of the structure, and place them in empty spots on the ground. Since all actions are reversible, a deconstruction episode played backward looks like the agent is building the goal structure. We consider the reversed episodes to be expert demonstrations. Programming an initialization function for each structure and a deconstruction policy is much easier than having a custom planner for each task. Note that the initialization function and deconstruction policy is used only for training tasks, not the testing task.

The objective of SDQfD is a weighted sum of a temporal-difference loss $L_\text{TD}$ (as in DQN \cite{mnih15}) and a term $L_\text{SLM}$ (with weight $\omega$) that penalizes action not selected by an expert with high predicted values 
\begin{align}
    L_\text{SDQfD} = L_\text{TD} + \omega \: L_\text{SLM}.
\end{align}

During training, the method identifies the set of non-expert actions $A_{s, a_e}$ with values higher than the value of an expert action $a_e$ minus a margin $l(a_e, a)$ (positive constant for $a_e \neq a$; zero otherwise)
\begin{align}
    A_{s, a_e} = \{ a \in A \:|\: Q(s, a) > Q(s, a_e)- l(a_e, a) \}.
\end{align}
The penalty called a "strict large margin" is then applied to the values of all actions in $A_{s, a_e}$
\begin{align}
    L_\text{SLM} = \frac{1}{|A_{s, a_e}|} \sum_{a \in A_{s, a_e}} Q(s, a)+ l(a_e, a) - Q(s, a_e).
\end{align}
We pre-train SDQfD on a replay buffer with expert demonstrations for 10k steps. After pre-training, we alternate between taking one environment step according to the current policy and performing a training step. We maintain two separate buffers: one for expert and one for on-policy transitions. Each batch of training data contains an equal number of samples from both buffers. The strict large margin is only computed for the expert data.

\section{Experiment Details}
\label{ap:sec:exp_details}

\subsection{Fruits World}
\label{ap:sec:exp_details:fruits}

We use the same neural network for both the DQN and the action prior: it has two hidden layers with 256 neurons and we use the ReLU non-linearity in-between. The DQN predicts a Q-value for each action and the action prior predicts the log probability of an action being optimal for each action separately.

Baseline DQNs are trained for 100k steps with an $\epsilon$-greedy policy that linearly decreases from 1.0 to 0.1 over 80k steps. The learning rate was set to $5 * 10^{-4}$ and we use prioritized replay with default parameters, dueling network and double Q-learning \cite{wang16,hasselt16,schaul16}.

Each action prior network is trained for 10k steps with a learning rate of 0.01. For the transfer experiments, we train a DQN with the same parameters as in the training tasks, except it uses an action prior for exploration, or it is initialized with Actor-Mimic weights (see Section \ref{sec:experiments:fruits}).

\subsection{Simulated Block Stacking}
\label{ap:sec:exp_details:blocks}

\begin{figure}[b!]
    \centering
    \includegraphics[width=1.0\textwidth]{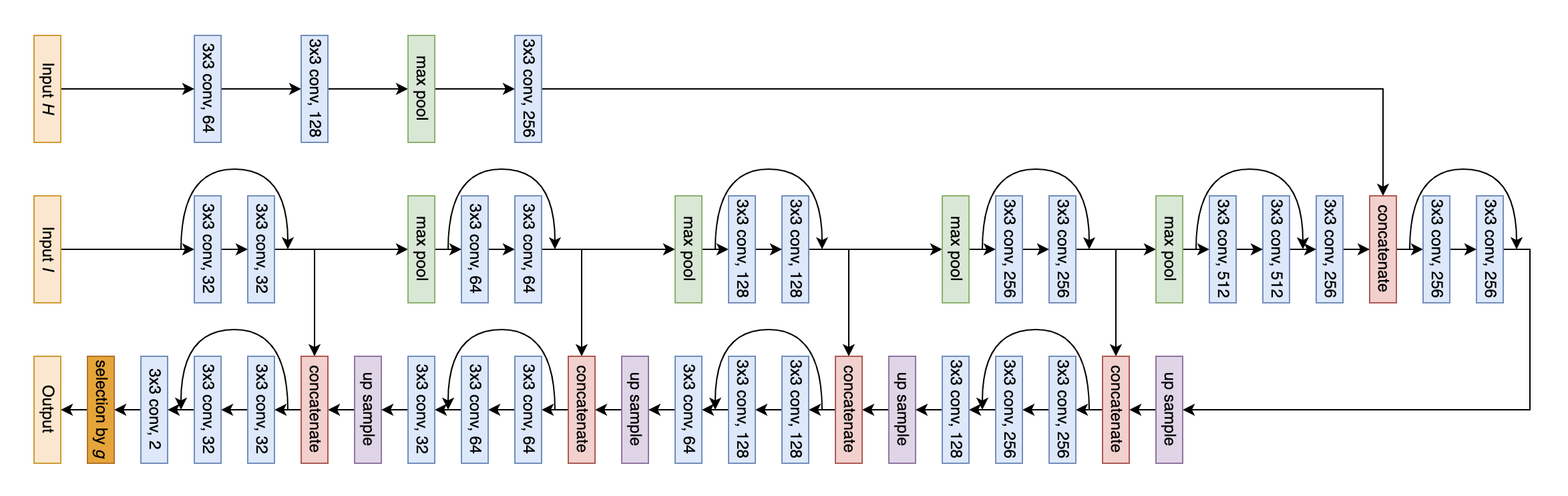}
    \caption{A modified U-Net network architecture. The inputs are a $90{\times}90$ depth image of the environment $I$ and a $24{\times}24$ zoomed-in image of an object the robot picked up. If the robot is not holding anything, all pixel values are set to zero. $3{\times}3$ conv, $32$ means a convolutional layer with 32 filters of size 3. All max pooling layers use a kernel of size 2 and a stride of 2. We omit the ReLU activations in the diagram.}
    \label{ap:fig:unet}
\end{figure}

We use a modified version of the U-Net architecture \cite{ronneberger15} for all our networks in this experiment. The schema of the fully-convolutional network is included in Figure \ref{ap:fig:unet}. 

To train the SDQfD, we collect 50k expert trajectories obtained by reversing deconstruction experience \cite{zakka20}. We pre-train model on this experience for 10k steps. Then we train the model while it interacts with the simulator for 40k episode. Each episode has a maximum length of 20. The learning rate is set to $5 * 10^{-5}$ and both the large margin weight and the margin coefficients are set to 0.1. There is no exploration, the batch size is set to 32 and the discount to 0.9. We run five simulated environments in parallel--we take one step in each environment, collect the transitions, take one training step of the SDQfD and repeat.

For the training datasets, we collect 20k steps for each of the 15 tasks used to train an action prior and concatenate the experience. The task classifier is trained on this data for 20k steps with a learning rate of $10^{-3}$, weight decay of $10^{-5}$ and batch size set to 32. We use a probability threshold for relevant tasks $\theta$ of 0.05. Action priors are trained with the same settings except for a batch size of 50 for 10k training steps.

In the transfer experiment, we use an action prior probability threshold $\sigma$ of 0.1. During exploration, the model only selects actions proposed by the action prior. We train a DQN for 100k episodes with the modified $\epsilon$-greedy policy. $\epsilon$ linearly decays from 1.0 to 0.01 for 80k episodes. The learning rate is set to $10^{-4}$, batch size to 32 and the discount factor to 0.9. We use prioritized experience replay with default parameters, but we do not use the weighting of the sampled transitions in the loss function.

In the weight sharing experiments, the weights $L^2$ penalty $\omega_{WS}$ is set to 0.1.

\subsection{Real-World Block Stacking}
\label{ap:sec:exp_details:blocks_real}
\begin{figure}[t]
\centering
\begin{subfigure}[t]{0.13\textwidth}
\centering
\includegraphics[width=1.0\textwidth]{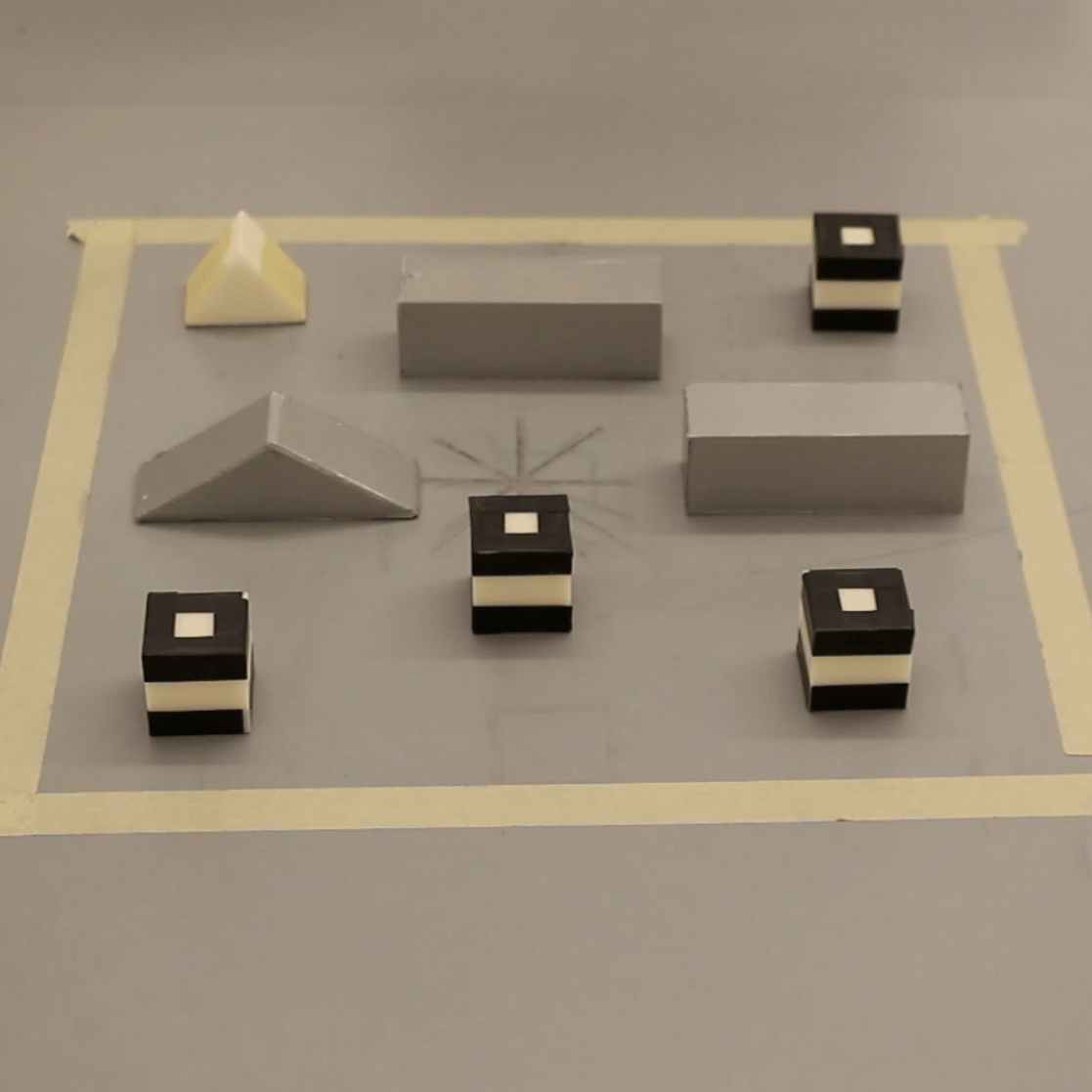}
\end{subfigure}
\begin{subfigure}[t]{0.13\textwidth}
\centering
\includegraphics[width=1.0\textwidth]{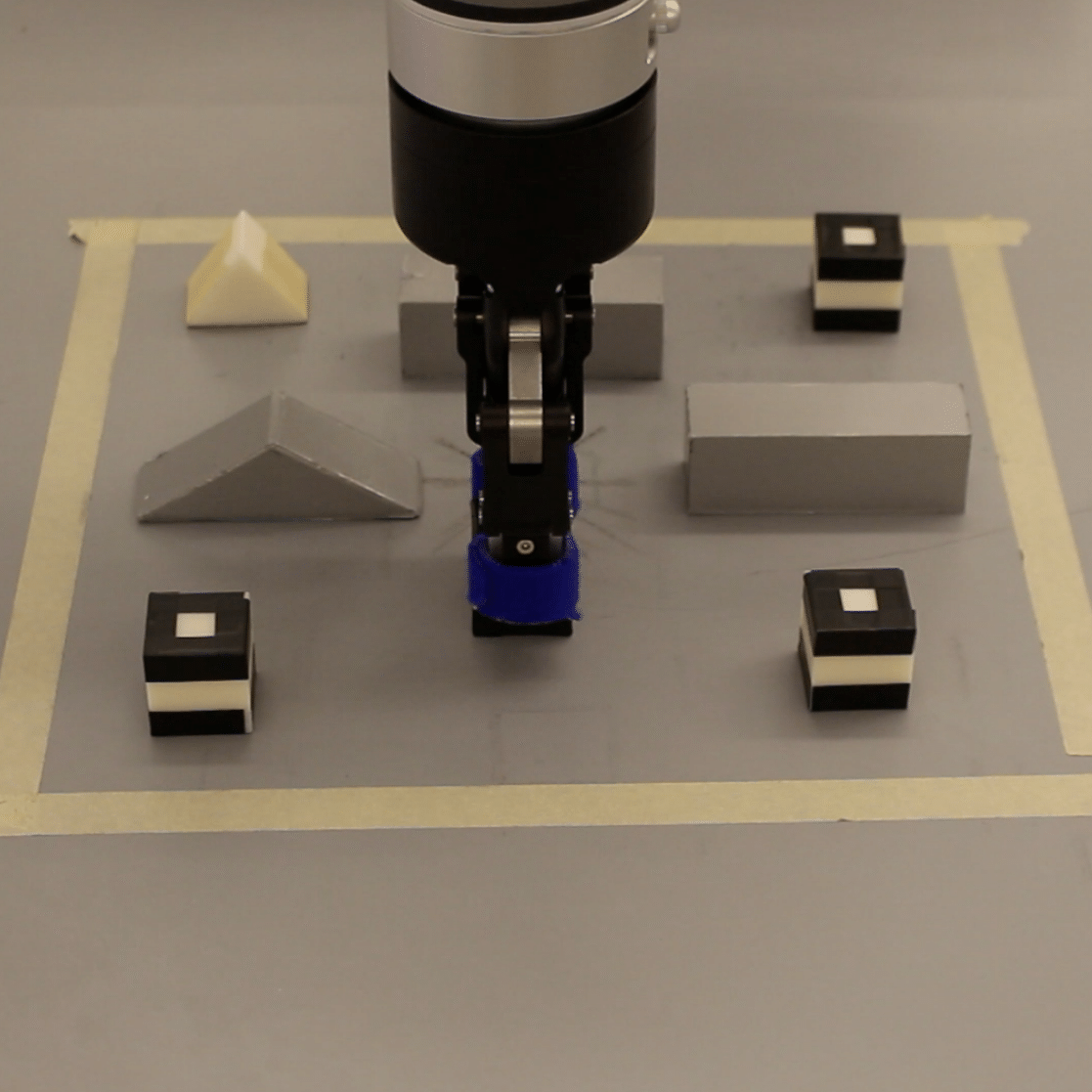}
\end{subfigure}
\begin{subfigure}[t]{0.13\textwidth}
\centering
\includegraphics[width=1.0\textwidth]{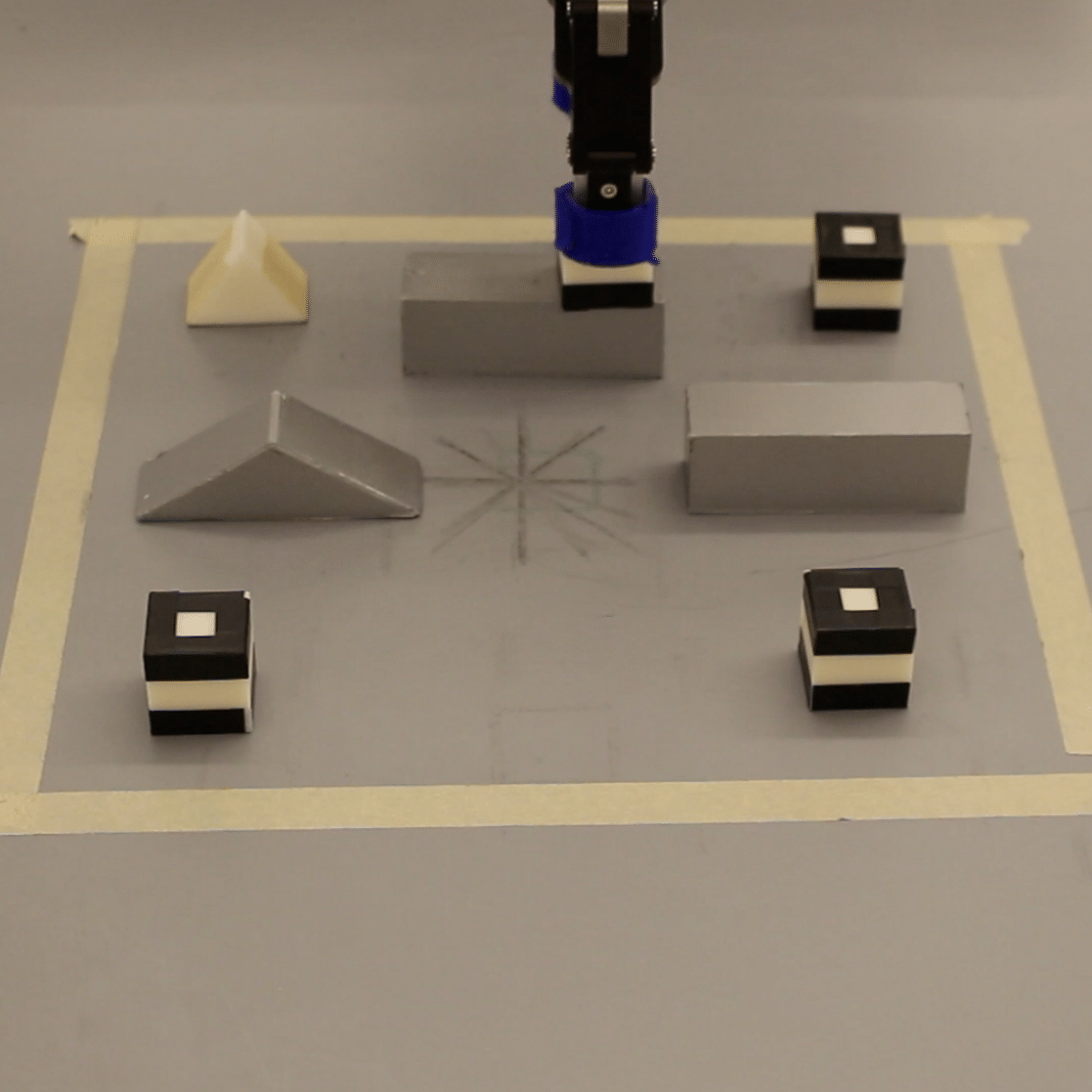}
\end{subfigure}
\begin{subfigure}[t]{0.13\textwidth}
\centering
\includegraphics[width=1.0\textwidth]{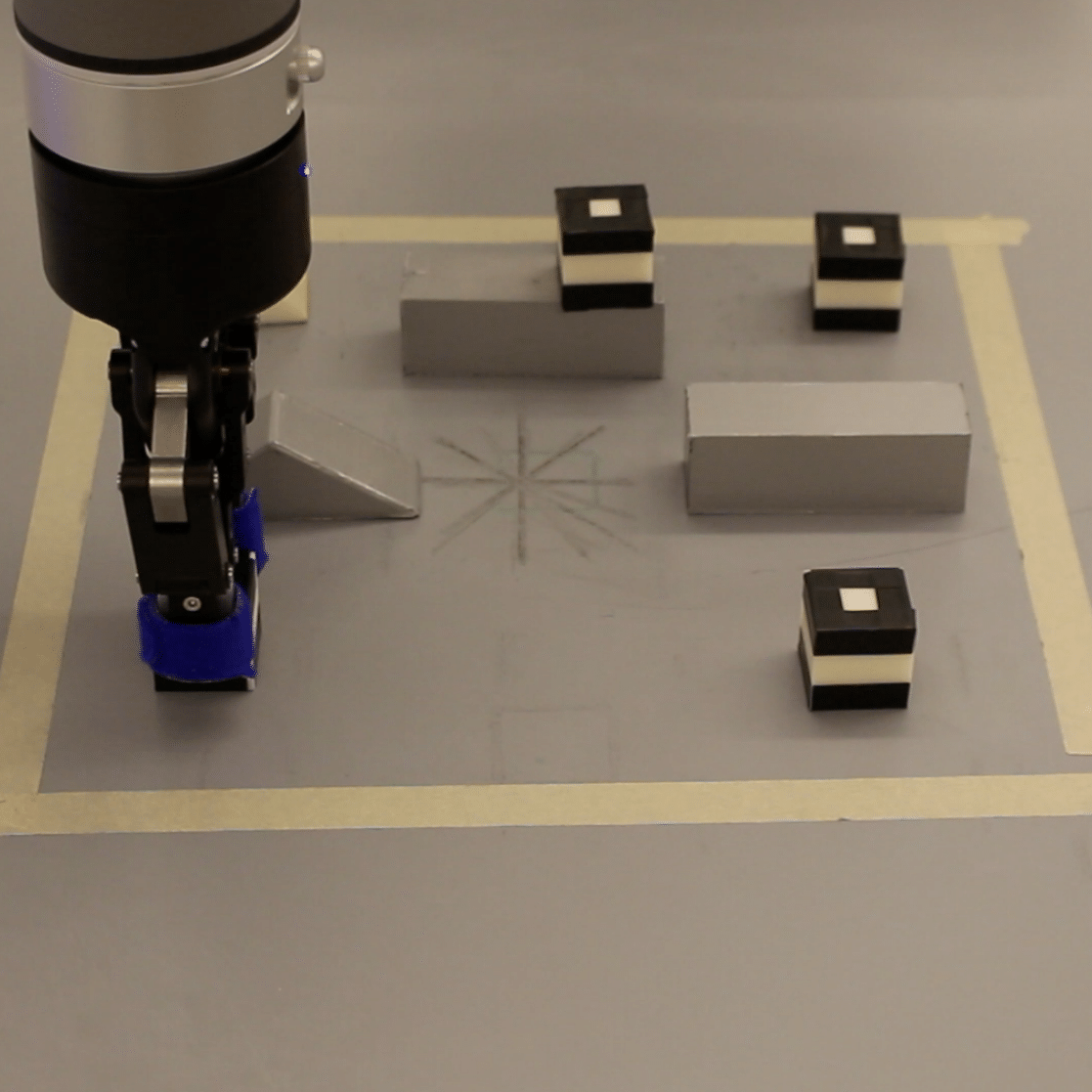}
\end{subfigure}
\begin{subfigure}[t]{0.13\textwidth}
\centering
\includegraphics[width=1.0\textwidth]{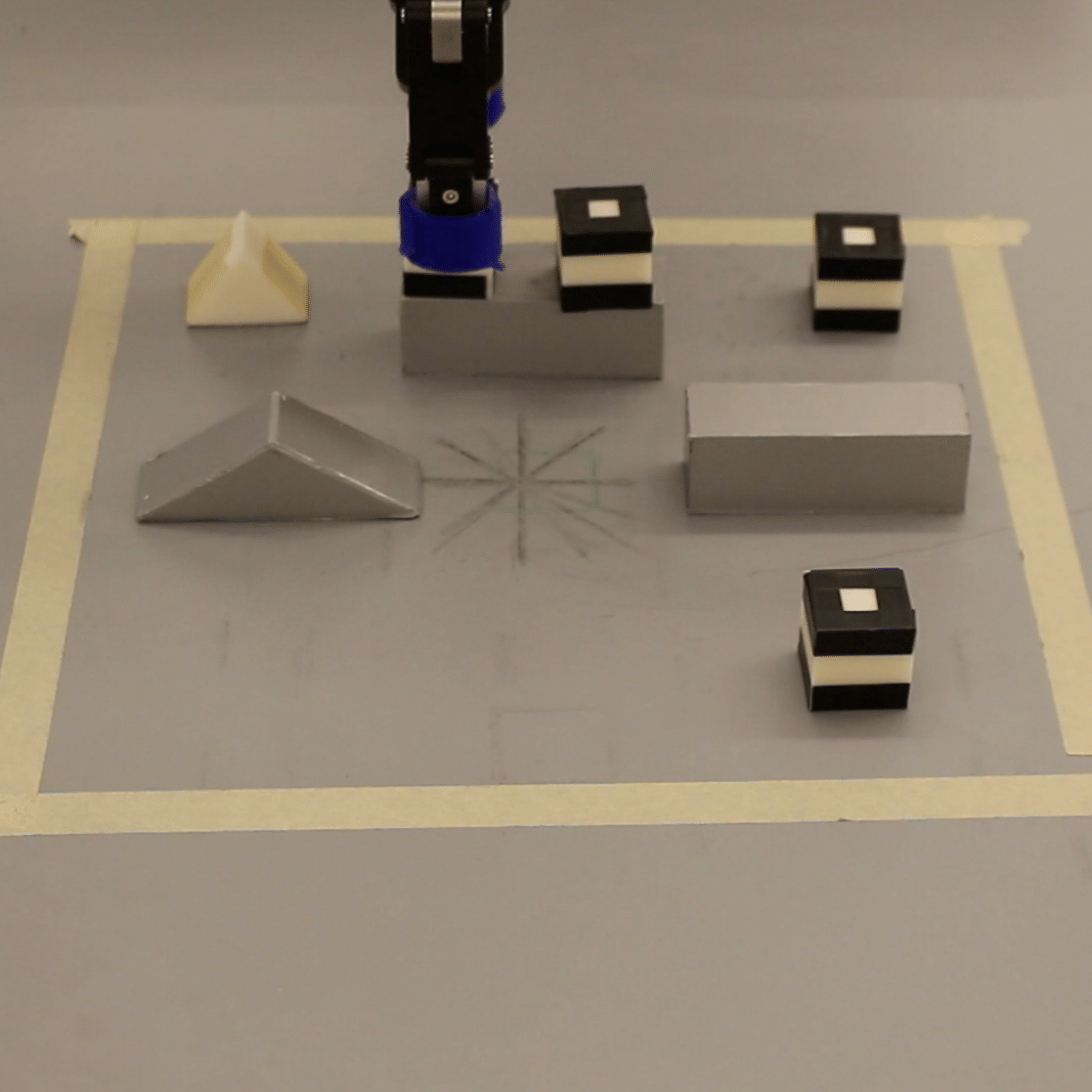}
\end{subfigure}
\begin{subfigure}[t]{0.13\textwidth}
\centering
\includegraphics[width=1.0\textwidth]{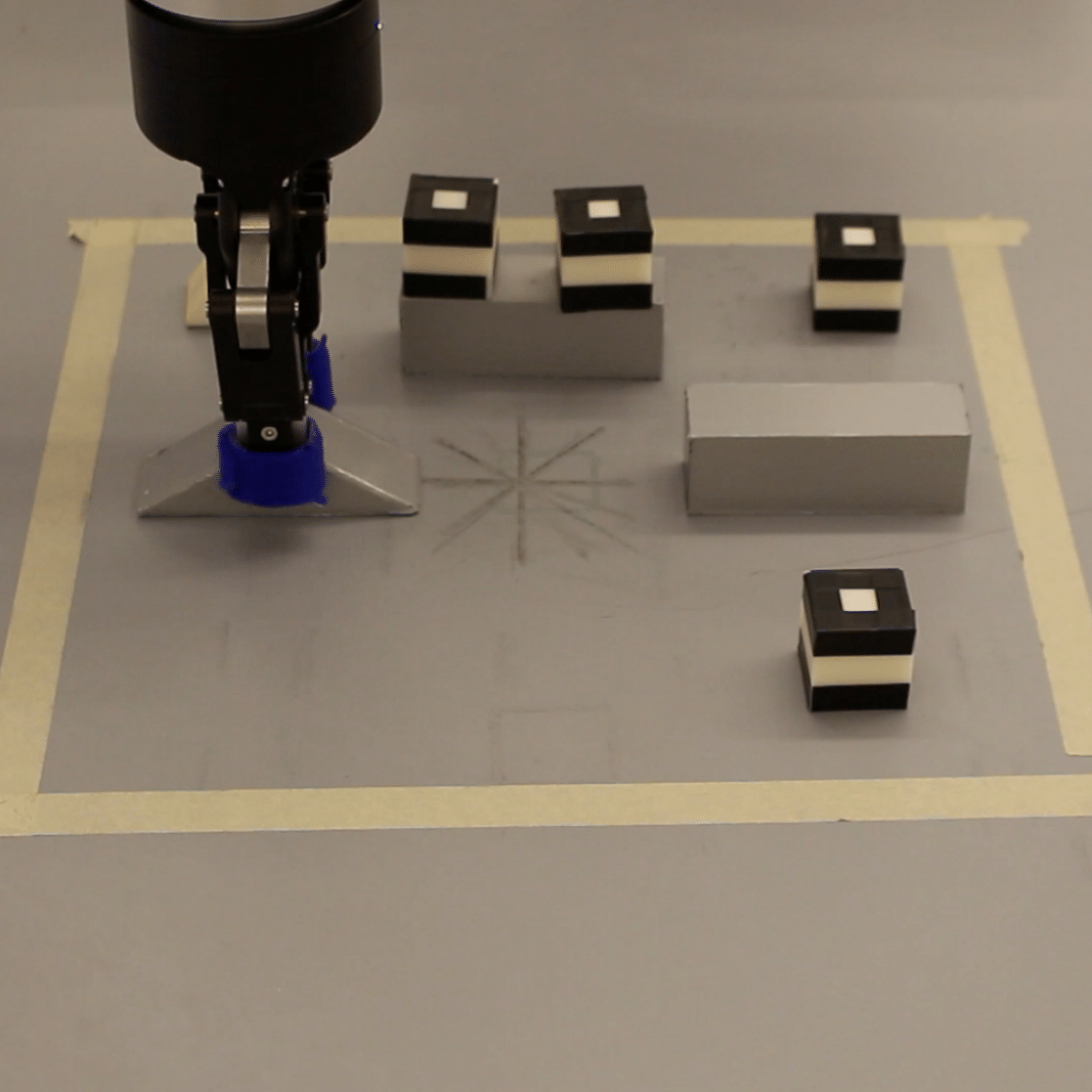}
\end{subfigure}
\begin{subfigure}[t]{0.13\textwidth}
\centering
\includegraphics[width=1.0\textwidth]{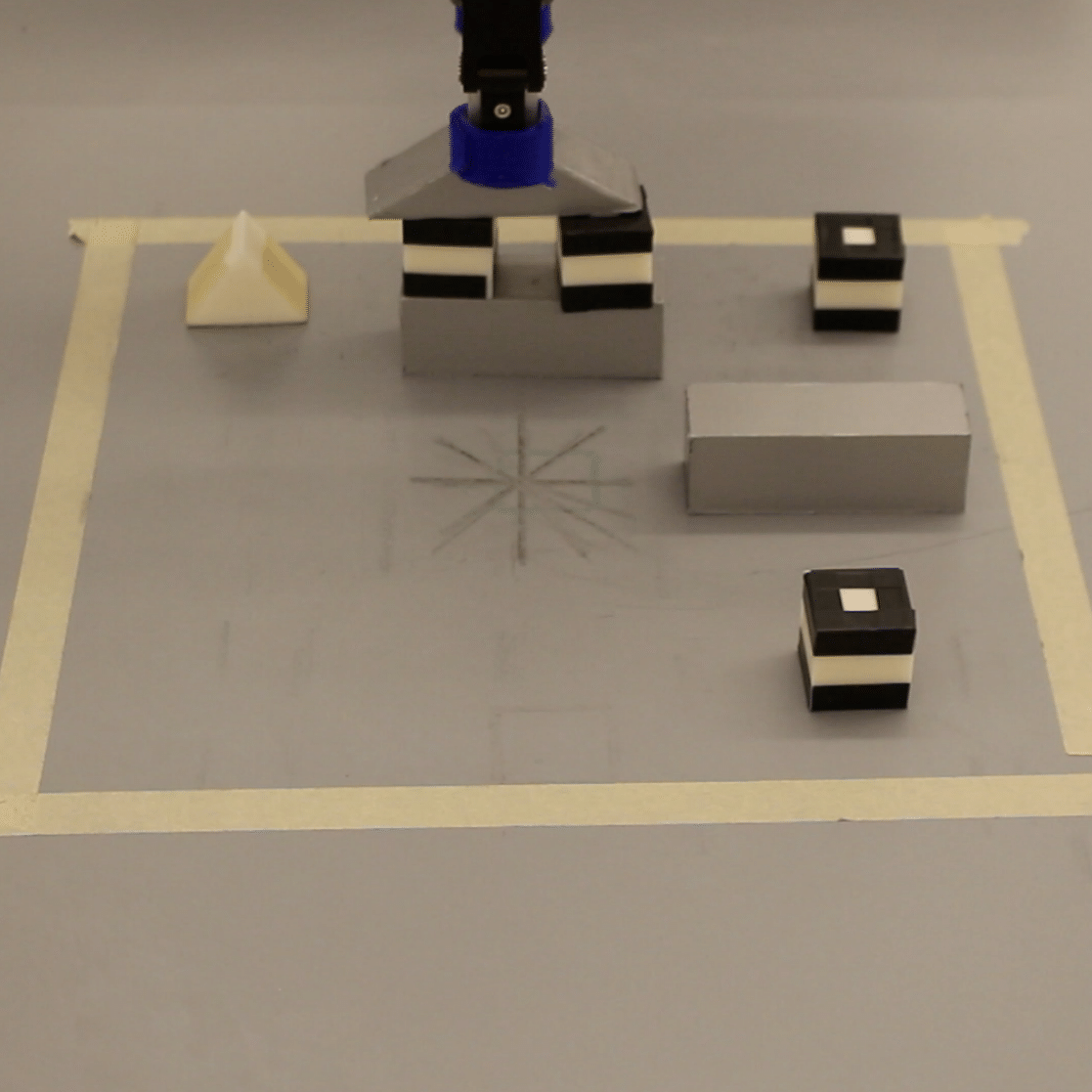}
\end{subfigure}
\caption{One example run of the 1l2b2r task.}
\label{ap:fig:robot_example}
\end{figure}

We tested the trained model on a Universal Robots UR5 arm with a Robotiq 2F-85 Gripper. The observation is provided by an Occipital Structure sensor pointing to the workspace from top-down. Figure \ref{fig:ur5} shows the robot experiment setup. All task parameters mirror the simulation. We run 10 trials for each task, and the maximum number of steps for each trial is 20. Figure \ref{ap:fig:robot_example} shows an example run of task 1l2b2r.

\section{Additional Results}

\begin{figure}[h!]
    \centering

    \begin{subfigure}[t]{0.24\textwidth}
        \centering
        \includegraphics[width=1.0\textwidth]{figures/exp/1b1r.pdf}
    \end{subfigure}
    \begin{subfigure}[t]{0.24\textwidth}
        \centering
        \includegraphics[width=1.0\textwidth]{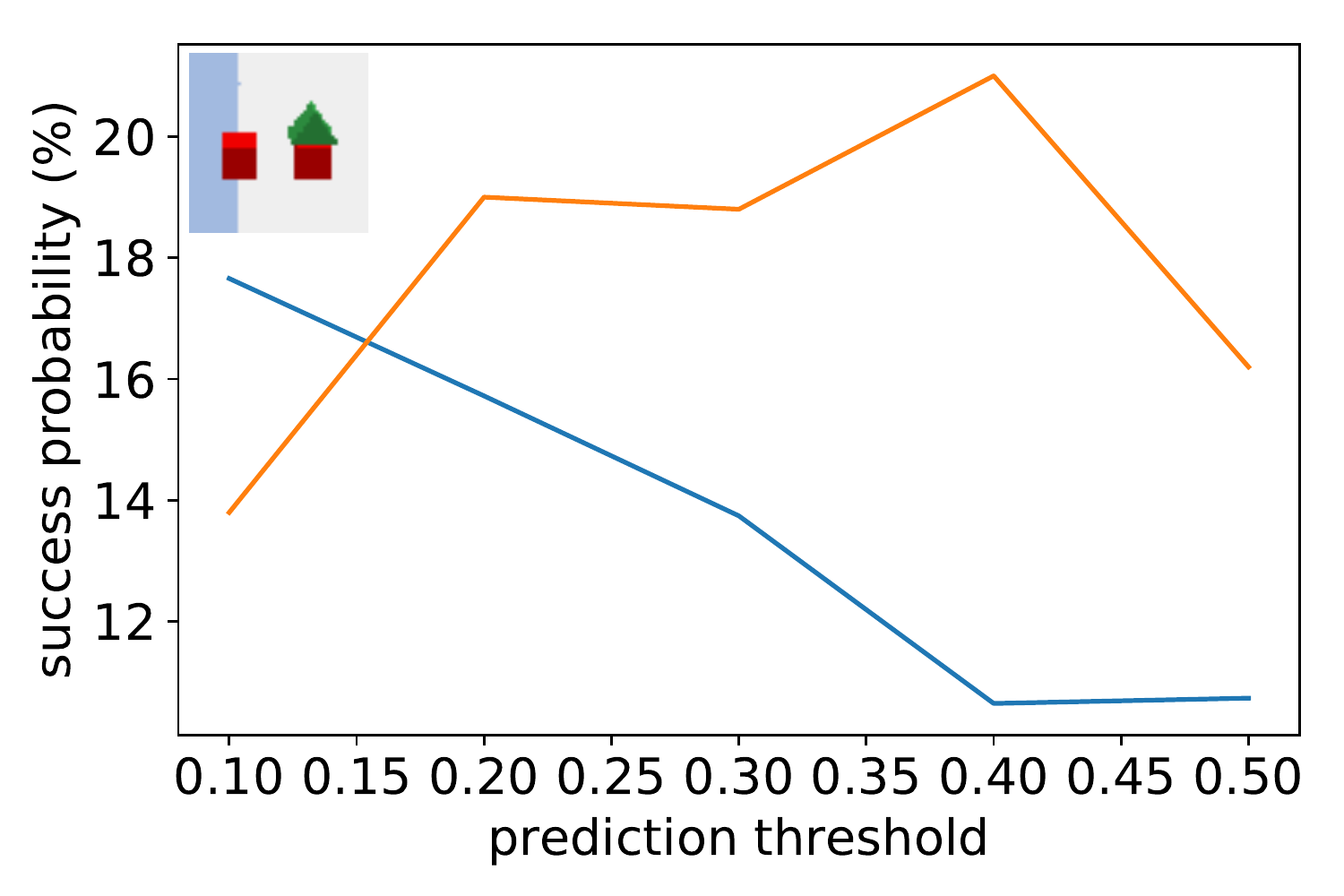}
    \end{subfigure}
    \begin{subfigure}[t]{0.24\textwidth}
        \centering
        \includegraphics[width=1.0\textwidth]{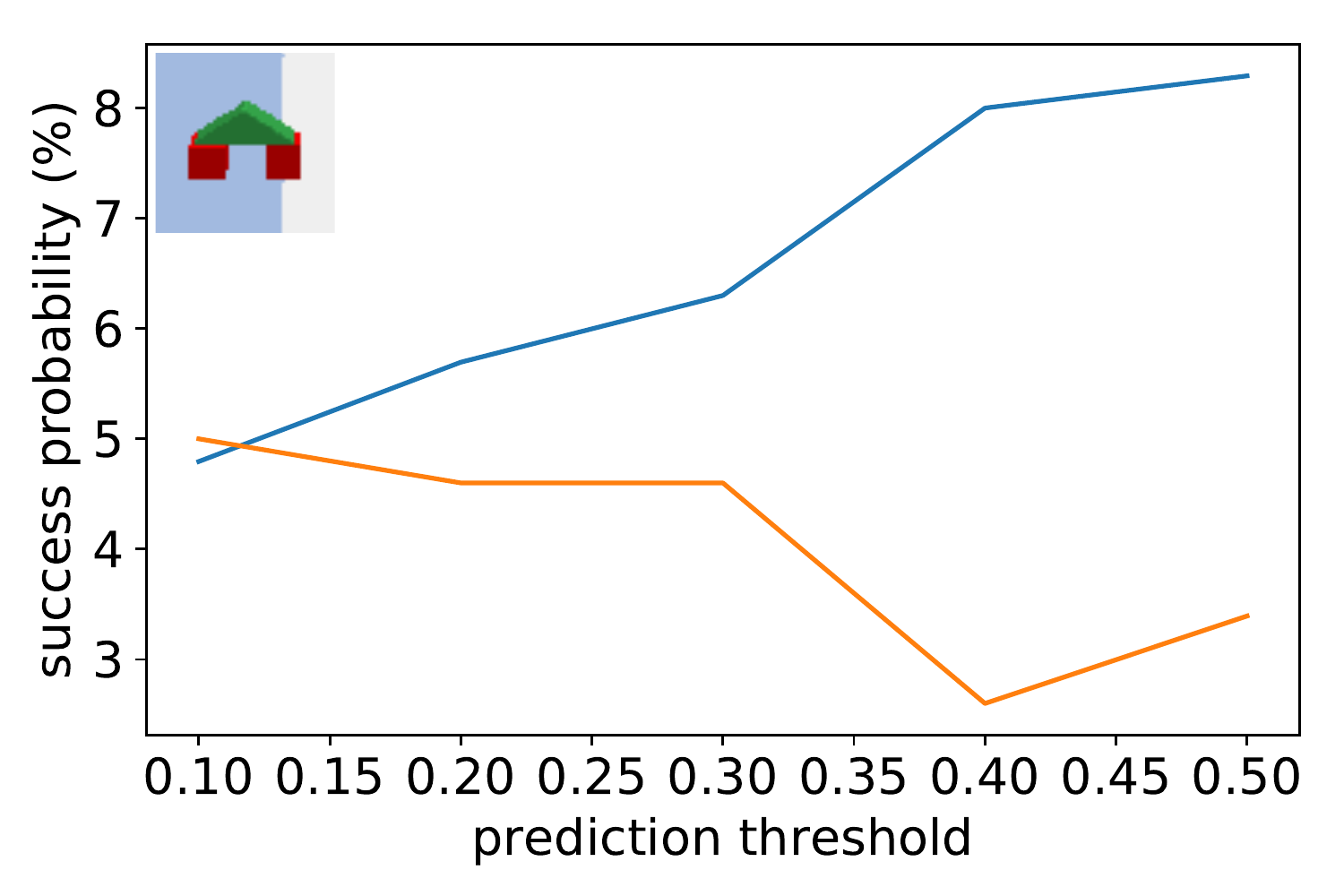}
    \end{subfigure}
    \begin{subfigure}[t]{0.24\textwidth}
        \centering
        \includegraphics[width=1.0\textwidth]{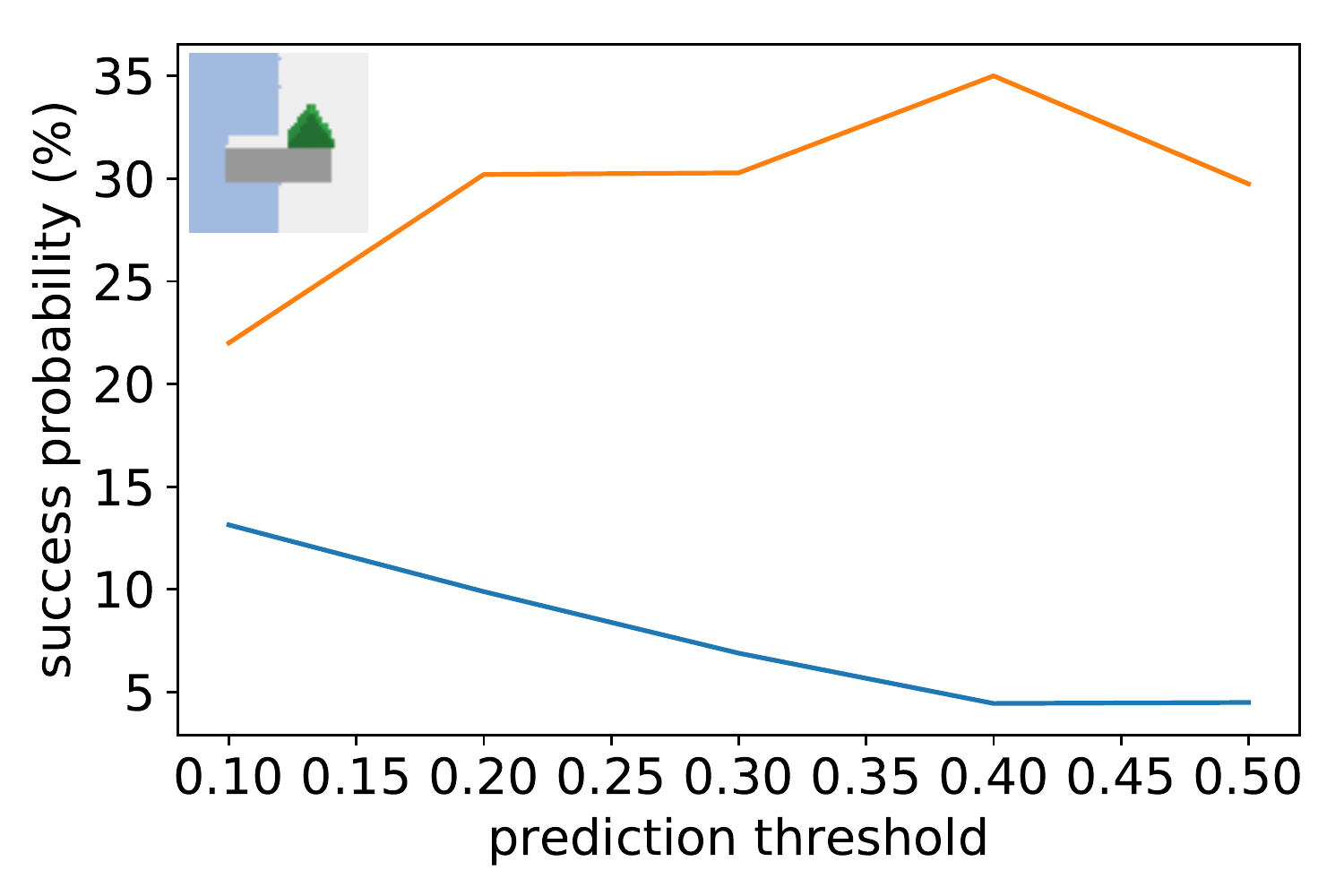}
    \end{subfigure}
    
    \begin{subfigure}[t]{0.24\textwidth}
        \centering
        \includegraphics[width=1.0\textwidth]{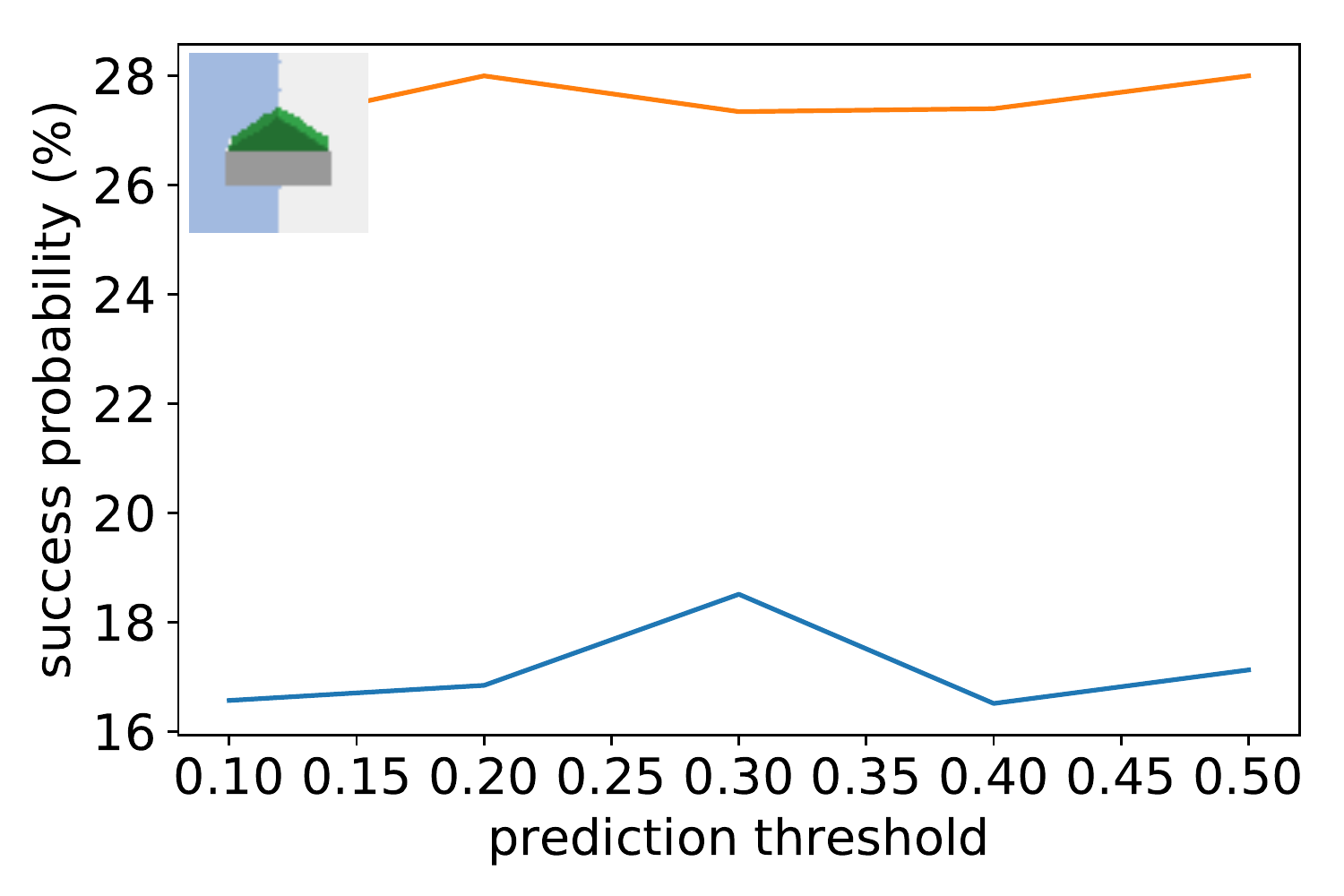}
    \end{subfigure}
    \begin{subfigure}[t]{0.24\textwidth}
        \centering
        \includegraphics[width=1.0\textwidth]{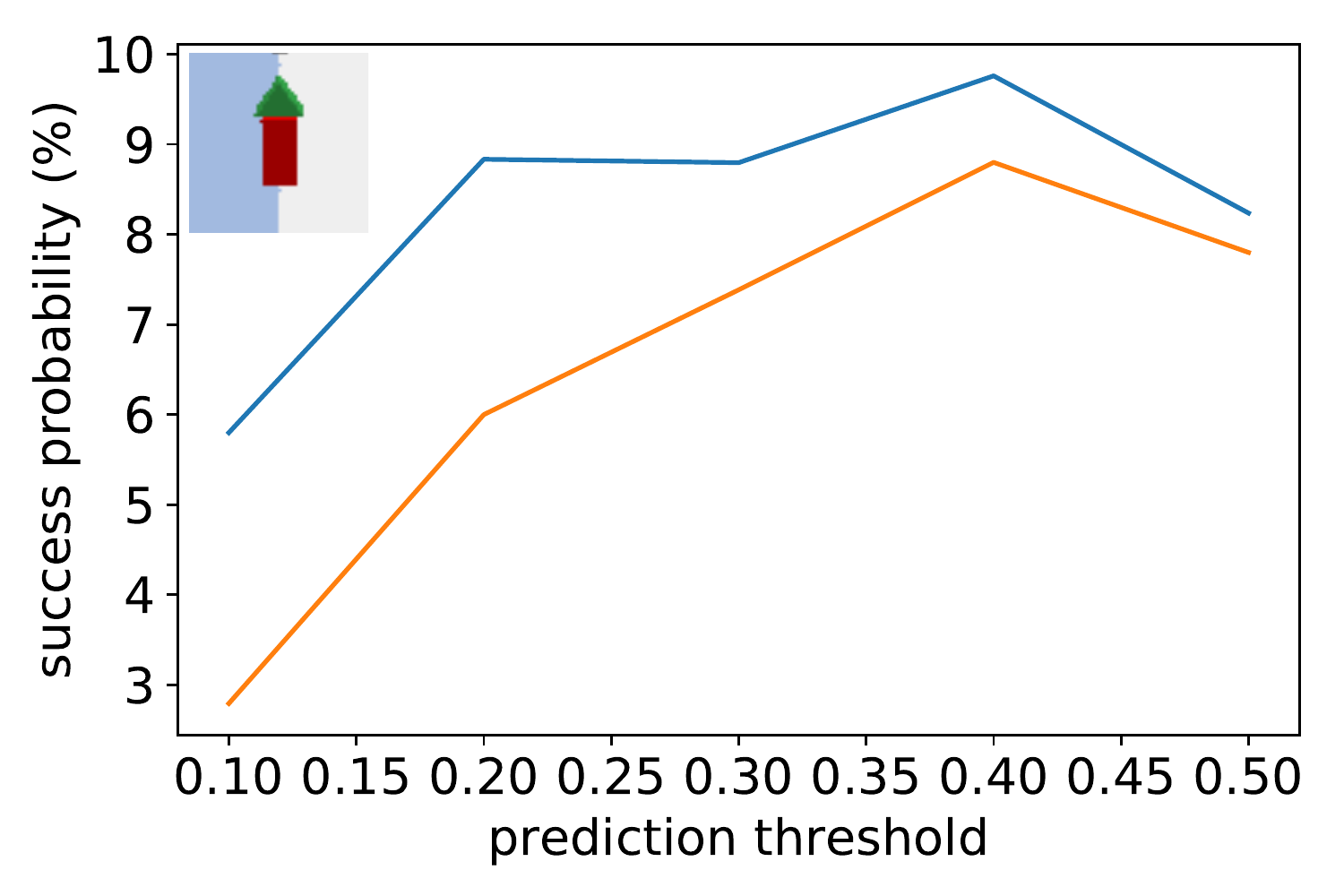}
    \end{subfigure}
    \begin{subfigure}[t]{0.24\textwidth}
        \centering
        \includegraphics[width=1.0\textwidth]{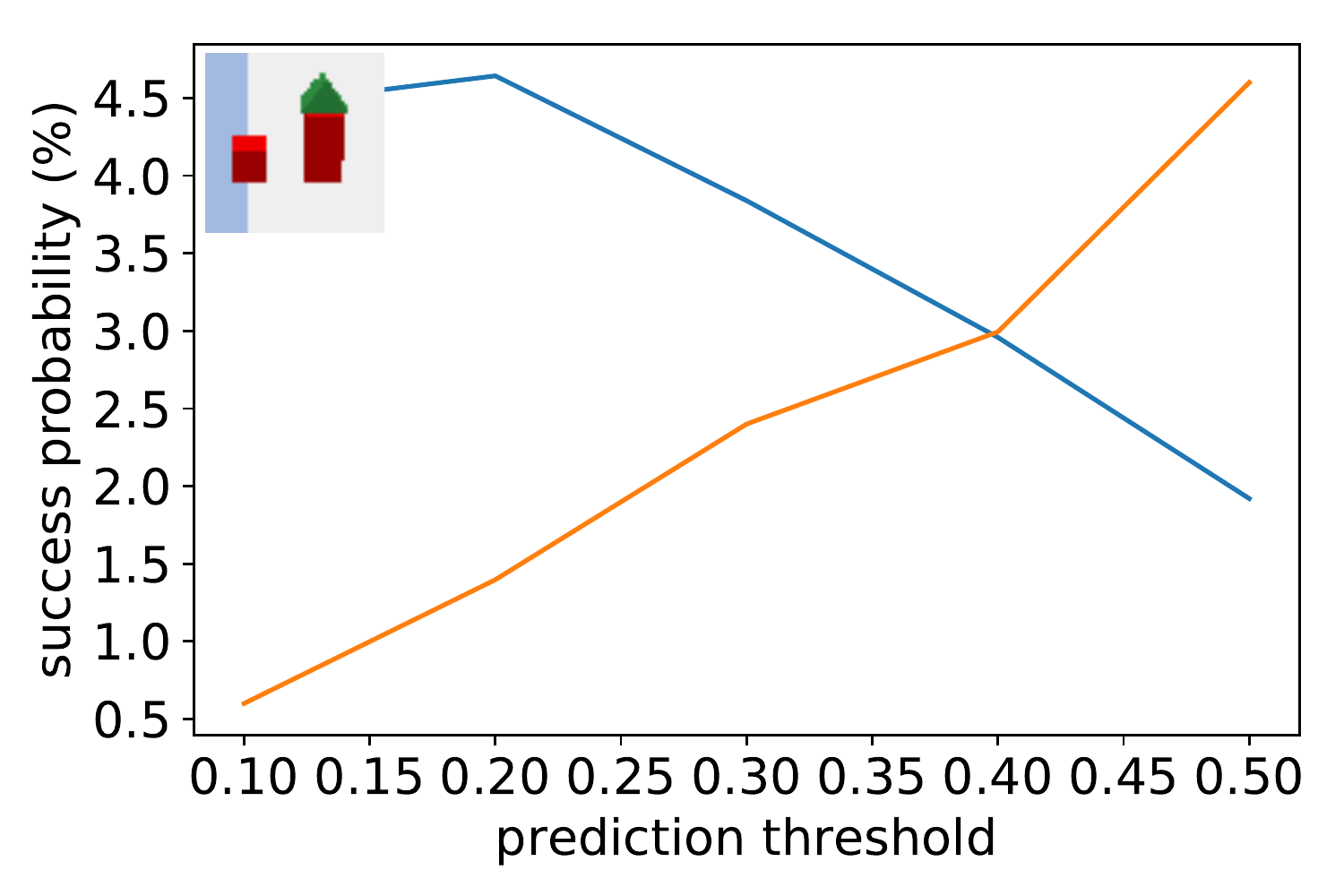}
    \end{subfigure}
    \begin{subfigure}[t]{0.24\textwidth}
        \centering
        \includegraphics[width=1.0\textwidth]{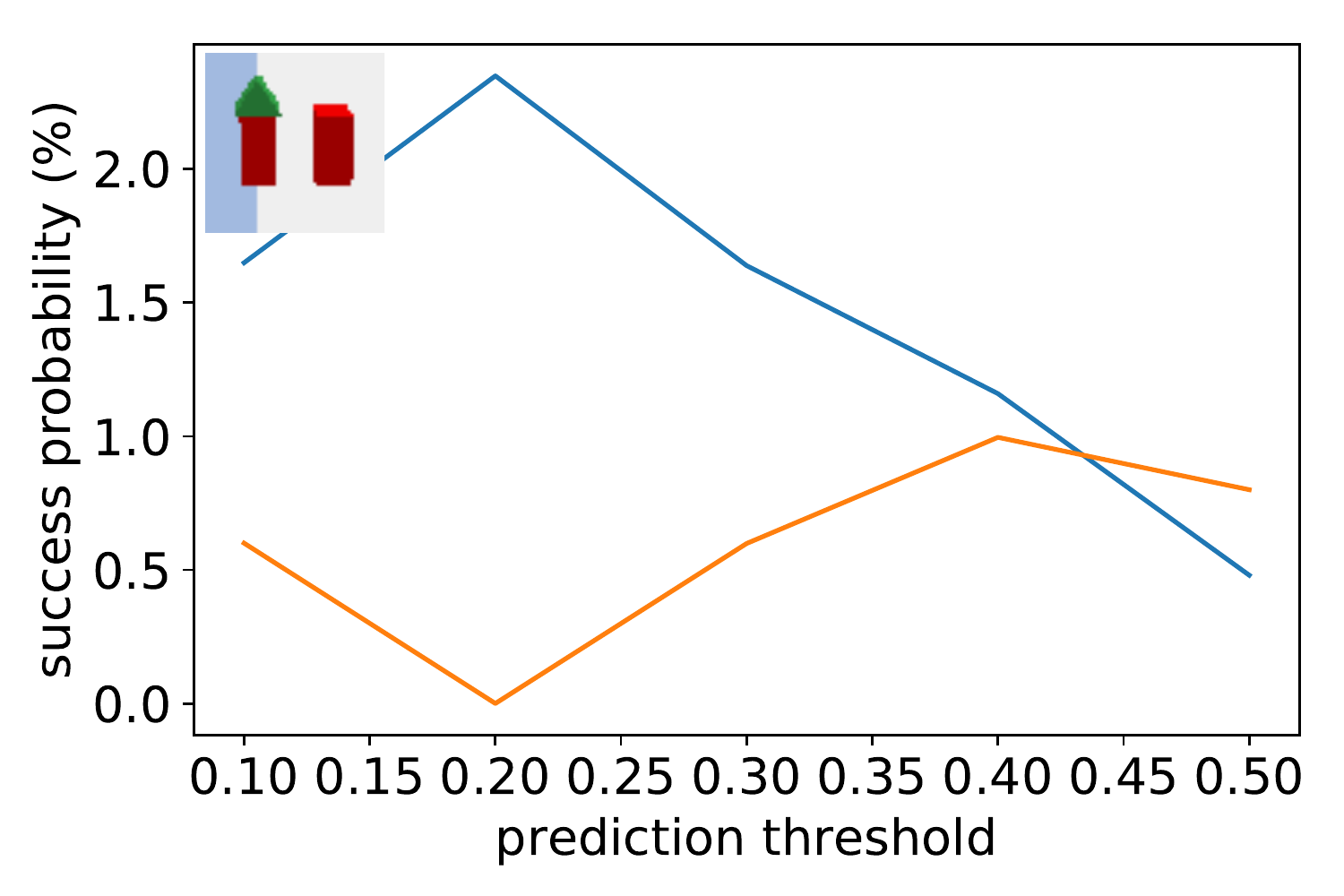}
    \end{subfigure}
    
    \begin{subfigure}[t]{0.24\textwidth}
        \centering
        \includegraphics[width=1.0\textwidth]{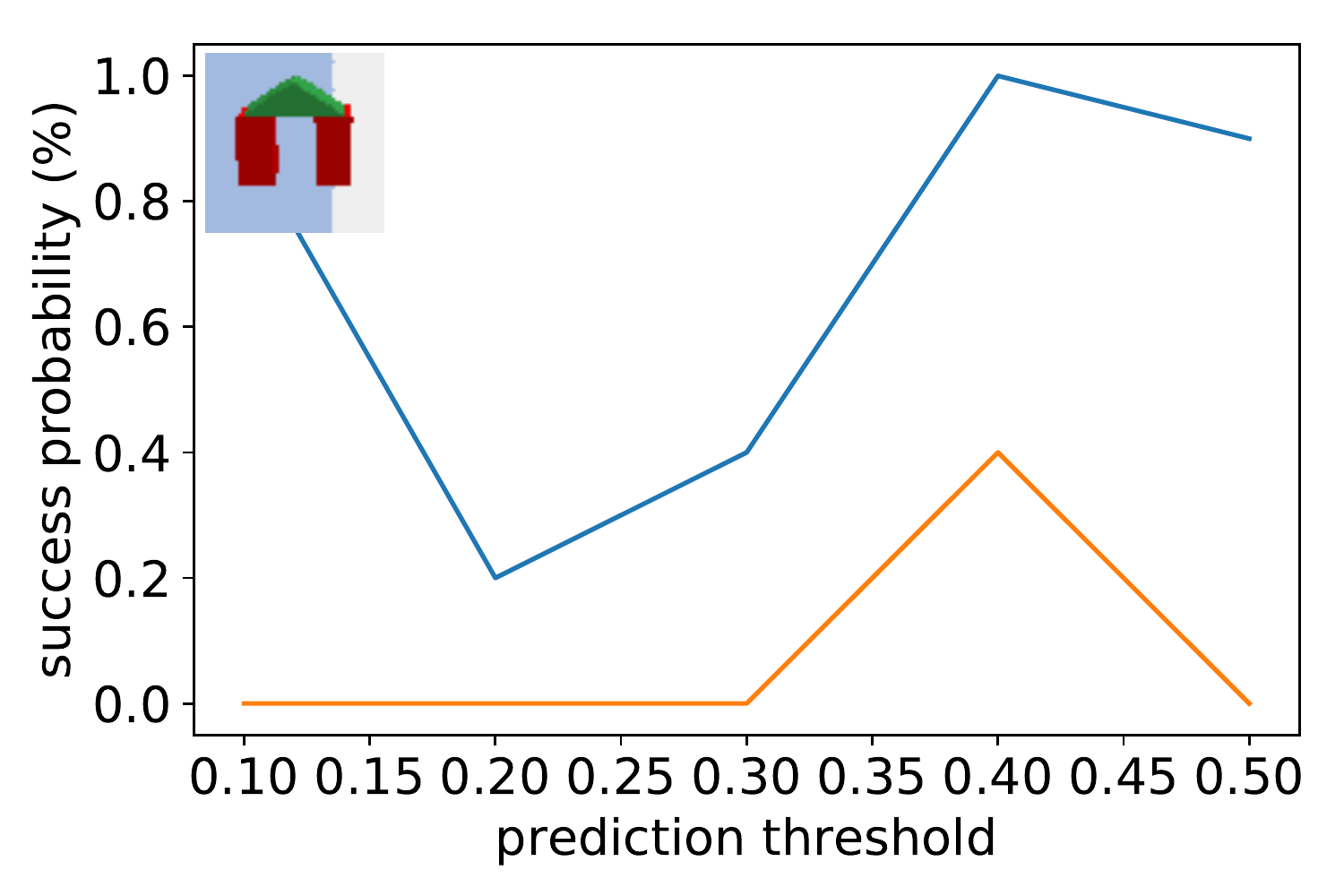}
    \end{subfigure}
    \begin{subfigure}[t]{0.24\textwidth}
        \centering
        \includegraphics[width=1.0\textwidth]{figures/exp/2b1l1r.pdf}
    \end{subfigure}
    \begin{subfigure}[t]{0.24\textwidth}
        \centering
        \includegraphics[width=1.0\textwidth]{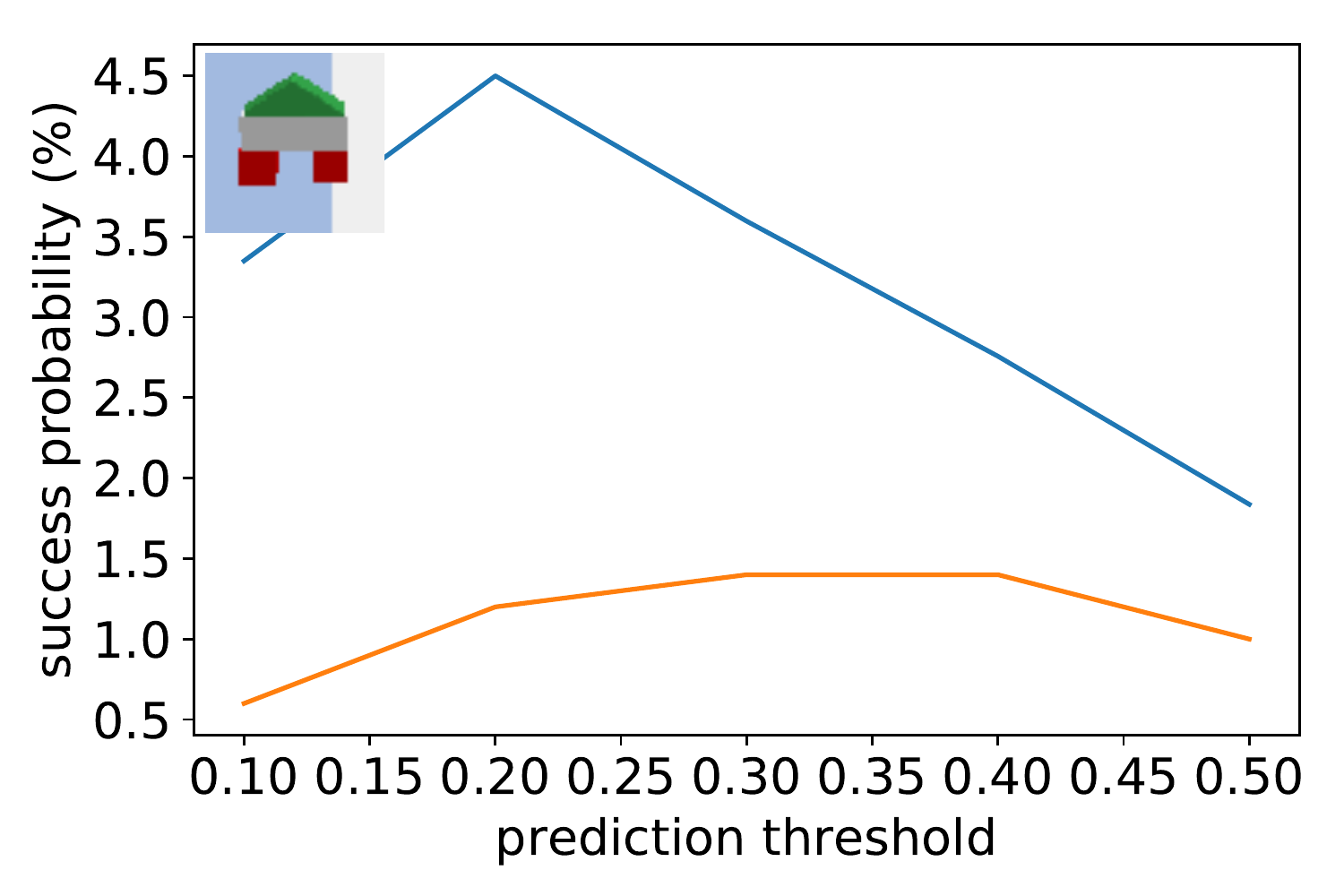}
    \end{subfigure}
    \begin{subfigure}[t]{0.24\textwidth}
        \centering
        \includegraphics[width=1.0\textwidth]{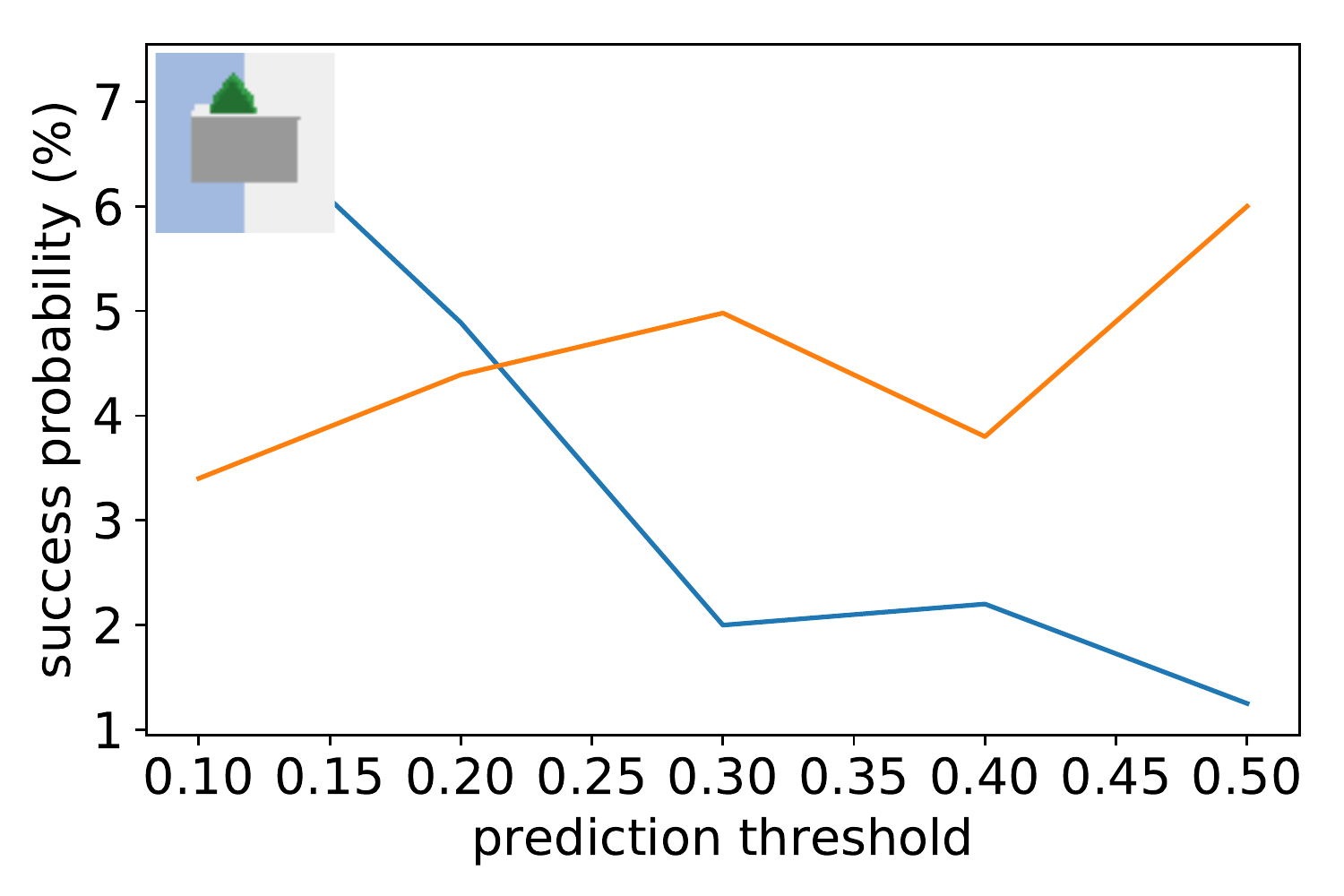}
    \end{subfigure}
    
    \begin{subfigure}[t]{0.24\textwidth}
        \centering
        \includegraphics[width=1.0\textwidth]{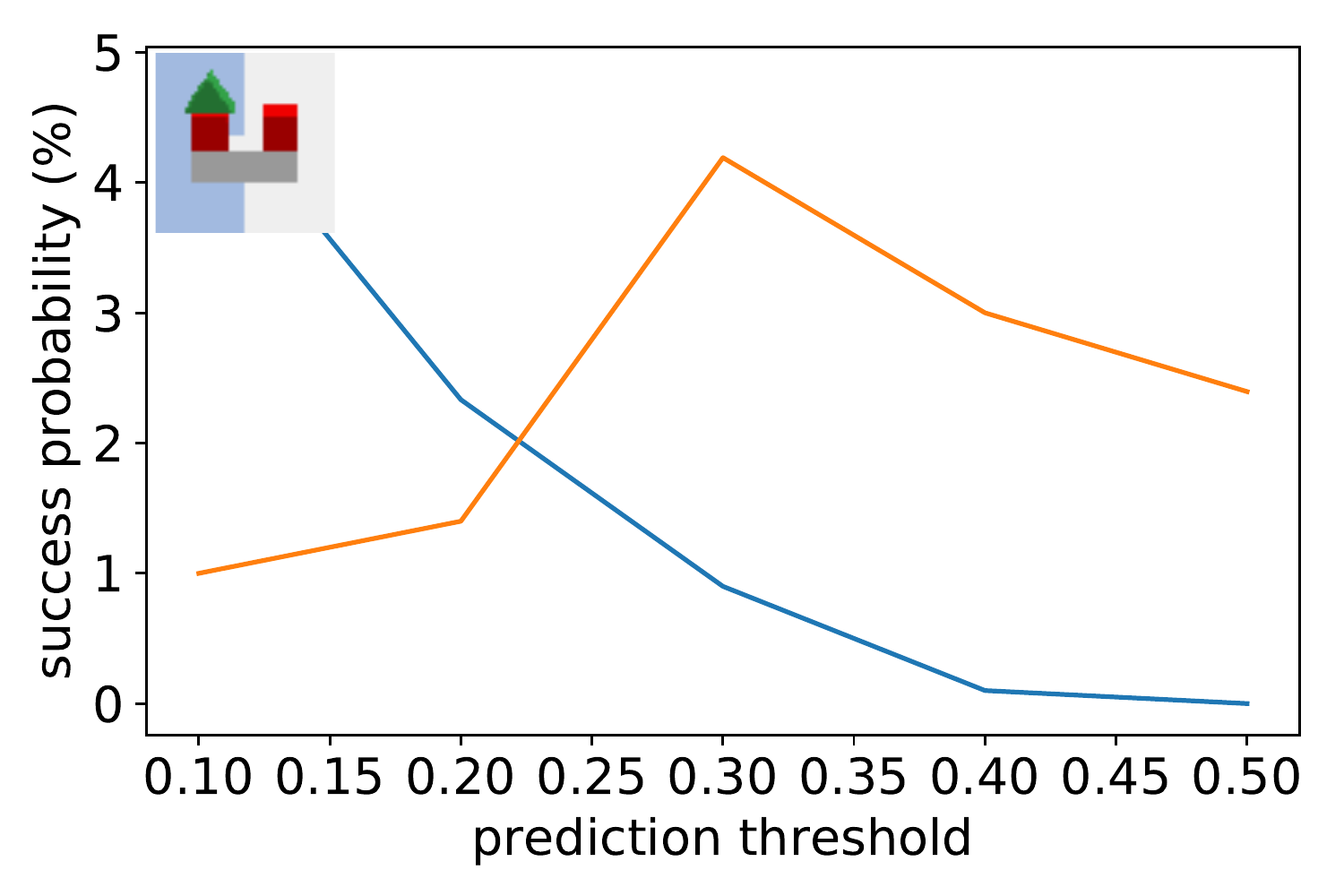}
    \end{subfigure}
    \begin{subfigure}[t]{0.24\textwidth}
        \centering
        \includegraphics[width=1.0\textwidth]{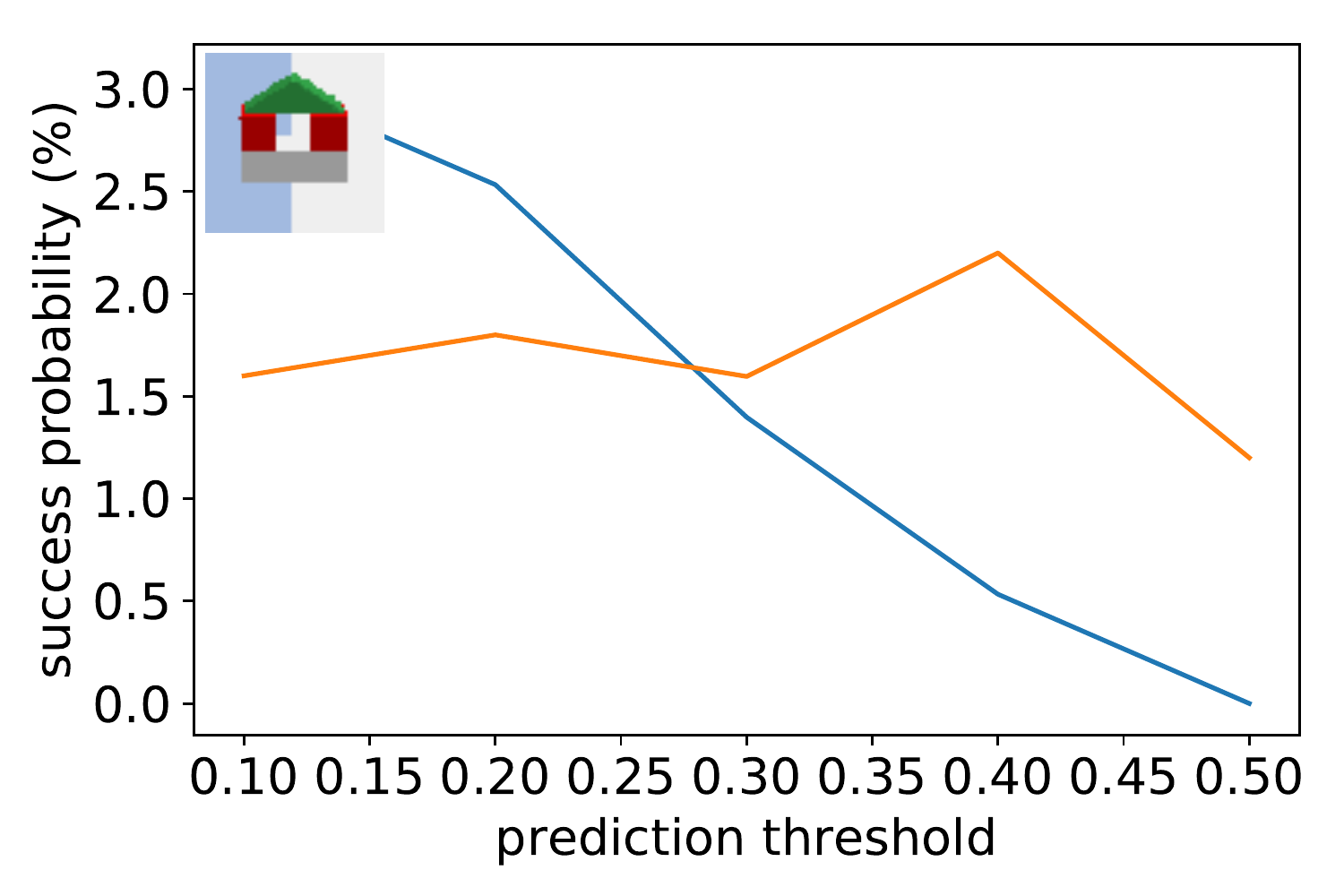}
    \end{subfigure}
    \begin{subfigure}[t]{0.24\textwidth}
        \centering
        \includegraphics[width=1.0\textwidth]{figures/exp/1l1l1r.pdf}
    \end{subfigure}
    \begin{subfigure}[t]{0.24\textwidth}
        \centering
        \includegraphics[width=1.0\textwidth]{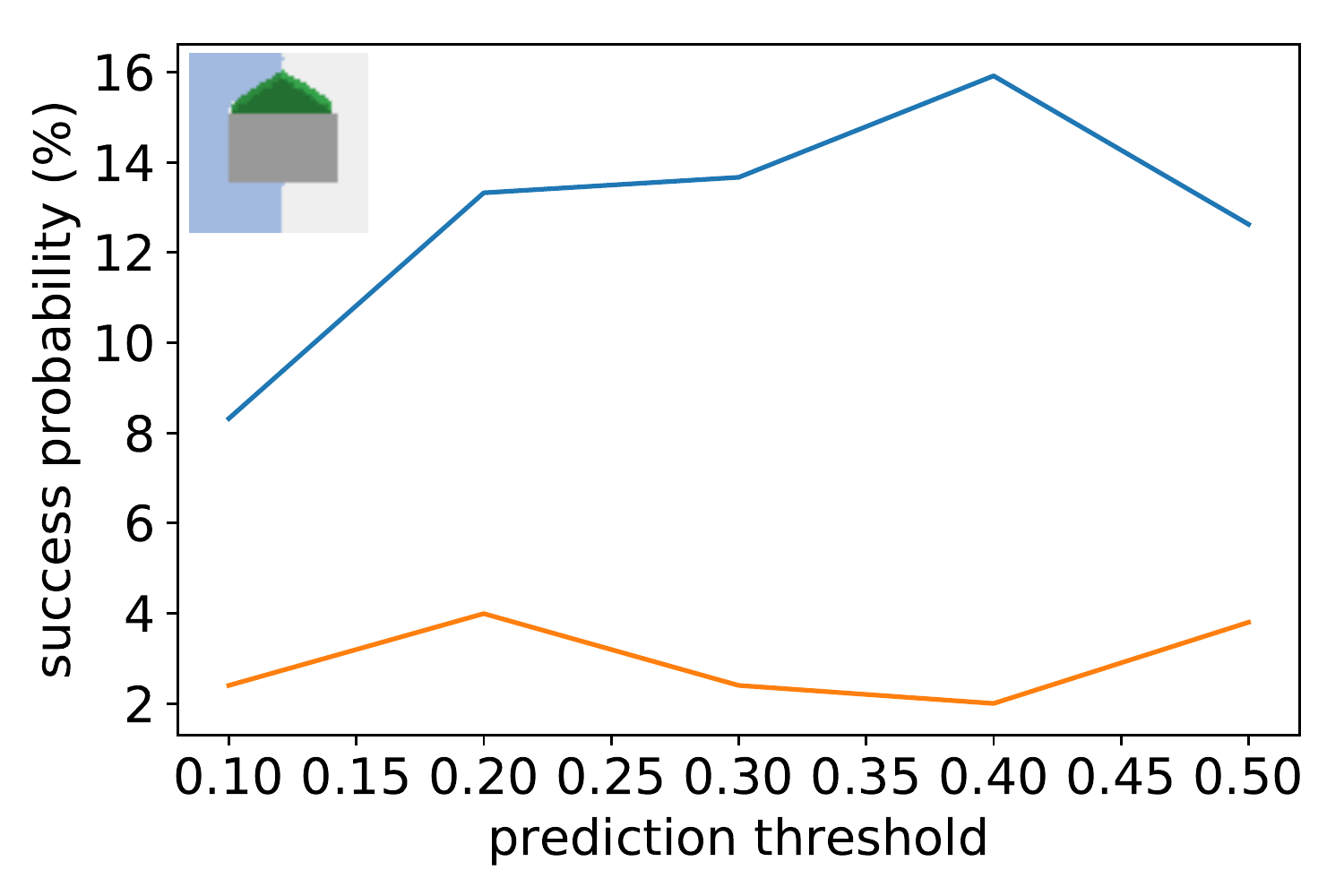}
    \end{subfigure}
 
    \caption{We show the same results as in Figure \ref{fig:block_exp} in the main test, but for all 16 tasks}
    \label{ap:fig:exp_all}
\end{figure}

Table \ref{ap:fig:exp_all} shows results for experiment \#2 (Section \ref{sec:experiments:blocks}) for all 16 tasks. As we stated in the main text, we found that task classifier helps in complex tasks and decreasing the probability threshold $\sigma$ tends to increase the success rates of the action priors that use the task classifier.

\begin{table}[t!]
    \centering
    \begin{tabular}{ccccccccccc}
        \toprule
        Num. fruits & \multicolumn{2}{c}{AP (ours)} & \multicolumn{2}{c}{DQN} & \multicolumn{2}{c}{AM-share} & \multicolumn{2}{c}{AM-freeze} & \multicolumn{2}{c}{AM-prog} \\
         & Mid. & Fin. & Mid. & Fin. & Mid. & Fin. & Mid. & Fin. & Mid. & Fin. \\
        \midrule
        \multicolumn{11}{c}{Fruit combinations} \\
        \midrule
        1 & 1 & 1 & 1 & 1 & 1 & 1 & 0.88 & 0.94 & 0.98 & 1 \\
        2 & 1 & 1 & 1 & 1 & 1 & 1 & 0.75 & 0.85 & 0.9 & 0.99 \\
        3 & 1 & 1 & 0.98 & 1 & 0.98 & 1 & 0.77 & 0.85 & 0.82 & 0.96 \\
        4 & 0.97 & 1 & 0.02 & 0.68 & 0.05 & 0.8 & 0.78 & 0.84 & 0.84 & 0.93 \\
        \midrule
        Mean & \textbf{0.99} & \textbf{1} & 0.75 & 0.92 & 0.76 & 0.95 & 0.8 & 0.87 & 0.89 & 0.97 \\
        \midrule
        \multicolumn{11}{c}{Fruit sequences} \\
        \midrule
        1 & 1 & 1 & 1 & 1 & 1 & 1 & 1 & 1 & 0.98 & 0.99 \\
        2 & 1 & 1 & 0.99 & 1 & 0.99 & 1 & 0.99 & 1 & 0.87 & 0.95 \\
        3 & 0.94 & 0.99 & 0.54 & 0.96 & 0.59 & 0.97 & 0.59 & 0.97 & 0.47 & 0.71 \\
        4 & 0.77 & 0.93 & -0.02 & 0 & -0.02 & 0 & -0.02 & 0 & -0.02 & -0.01 \\
        \midrule
        Mean & \textbf{0.93} & \textbf{0.98} & 0.63 & 0.74 & 0.64 & 0.74 & 0.64 & 0.74 & 0.58 & 0.66 \\
        \bottomrule
    \end{tabular}
    \vspace{0.5em}
    \caption{Additional Fruits World results accompanying Figure \ref{fig:fruits_learning_curves}. We report the average reward over 10 runs in the middle of training (Mid.) and at the end (Fin.). Num. fruits refers to the number of fruits the agent is supposed to pick--it determines the difficulty of the task. Each row is an average over all tasks involving picking up the specified number of fruits.}
    \label{tab:fruits_seq}
\end{table}


\end{document}